\documentclass[lettersize,journal]{IEEEtran}
\usepackage{bm}
\usepackage{amsmath,amsfonts}
\usepackage{algorithmic}
\usepackage{algorithm}
\usepackage{array}
\usepackage[caption=false,font=normalsize,labelfont=sf,textfont=sf]{subfig}
\usepackage{textcomp}
\usepackage{stfloats}
\usepackage{url}
\usepackage{verbatim}
\usepackage{graphicx}
\usepackage{xcolor}
\usepackage{cite}
\usepackage{enumitem}
\usepackage{multirow}
\usepackage{array}
\usepackage{CJKutf8}
\usepackage{threeparttable}
\usepackage{tikz}
\usepackage[edges]{forest}
\usepackage[colorlinks,linkcolor=blue]{hyperref}
\usepackage{adjustbox}
\usepackage{makecell}

\begin{document}



\title{A Survey on Medical Large Language Models: Technology, Application, Trustworthiness, and Future Directions}

\author{Lei Liu, Xiaoyan Yang, Junchi Lei, Yue Shen, Jian Wang, Peng Wei, Zhixuan Chu, Zhan Qin, Kui Ren 
\thanks{Corresponding Authors: Zhixuan Chu (zhixuanchu@zju.edu.cn)}
\thanks{The authors (Xiaoyan Yang, Yue Shen, Jian Wang, and Peng Wei) are with Ant Group. The authors (Junchi Lei, Zhixuan Chu, Zhan Qin, Kui Ren) are with Zhejiang University. Lei Liu is with The Chinese University of Hong Kong, Shenzhen, Guangdong, China. }
\thanks{Lei Liu, Xiaoyan Yang, and Junchi Lei contribute equally to this work. Work was done during Lei's Internship at Ant Group (liulei1497@gmail.com).}
}

\markboth{Journal of \LaTeX\ Class Files,~Vol.~14, No.~8, August~2024}%
{Shell \MakeLowercase{\textit{et al.}}: A Sample Article Using IEEEtran.cls for IEEE Journals}

\IEEEpubid{0000--0000/00\$00.00~\copyright~2024 IEEE}

\maketitle

\begin{abstract}
With the advent of Large Language Models (LLMs), medical artificial intelligence (AI) has experienced substantial technological progress and paradigm shifts, highlighting the potential of LLMs to streamline healthcare delivery and improve patient outcomes. Considering this rapid technical progress, in this survey, we trace the recent advances of Medical Large Language Models (Med-LLMs), including the background, key findings, and mainstream techniques, especially for the evolution from general-purpose models to medical-specialized applications. Firstly, we delve into the foundational technology of Med-LLMs, indicating how general models can be progressively adapted and refined for the complicated medical tasks. Secondly, the wide-ranging applications of Med-LLMs are investigated across various healthcare domains, as well as an up-to-date review of existing Med-LLMs. The transformative impact of these models on daily medical practice is evident through their ability to assist clinicians, educators, and patients. Recognizing the importance of responsible innovation, we discuss the challenges associated with ensuring fairness, accountability, privacy, and robustness. Ethical considerations, rigorous evaluation methodologies, and the establishment of regulatory frameworks are crucial for building trustworthiness in the real-world system. We emphasize the need for ongoing scrutiny and development to maintain high standards of safety and reliability. Finally, we anticipate possible future trajectories for Med-LLMs, identifying key avenues for prudent expansion. By consolidating these insights, our review aims to provide professionals and researchers with a thorough understanding of the strengths and limitations of Med-LLMs, fostering a balanced and ethical approach to their integration into the healthcare ecosystem.
\end{abstract}


\section{Introduction}
The emergence of foundation models has sparked a transformative wave within the Artificial Intelligence (AI) community in recent years. The success, particularly their remarkable generalization capacity across a myriad of downstream tasks, is attributed to significantly large model sizes and large-scale pre-training on expansive datasets \cite{xue2023prompt,chu2023data}. Such progression has witnessed the extraordinary evolution of large language models (LLMs) within the Natural Language Processing (NLP) society, characterized by a series of seminal advancements, such as BERT \cite{devlin2018bert}, T5 \cite{2020t5}, and the latest GPT models \cite{gpt4all}. Leveraging these powerful models, AI systems can now excel in generating human-like responses, hinting at the potential for highly sophisticated and interactive applications.

The prevail of LLMs creates an unprecedented opportunity for AI technologies to make significant contributions to the medical domain, \textit{i.e.}, Medical Large Language Models (Med-LLMs). More recently, the exploration of Med-LLMs covers from helping clinicians make more accurate decisions to improving patient care quality and outcomes, such as ChiMed-GPT~\cite{tian2023chimed}, MedicalGPT~\cite{MedicalGPT}, HuatuoGPT-II~\cite{chen2023huatuogpt}, and ChatMed~\cite{zhu2023ChatMed}, which have garnered increasing interests within both academic and industry communities due to their potential to transform various aspects of healthcare and biomedical applications. Med-LLMs could bring a multitude of advantages to healthcare, including enhanced medical knowledge comprehension, improved diagnostic accuracy, personalized treatment recommendations, \textit{etc}. For instance, MedPrompt \cite{nori2023can} can achieve superior results on the United States Medical Licensing Examination (USMLE), outperforming expert-level human \textit{i.e.}, 90.2 \textit{VS.} 87.0. HuatuoGPT-II~\cite{chen2023huatuogpt} can successfully pass multiple Chinese medical qualification examinations.

\IEEEpubidadjcol

Med-LLMs pave the way for more sophisticated, adaptable, and trustworthy clinical workflows in healthcare. Before the advent of Med-LLMs, researchers predominantly pay more efforts to adapt pre-trained language models (PLMs) for clinical applications \cite{min2023recent}, which are equipped with relatively small model sizes (\textit{e.g.}, BERT~\cite{devlin2018bert} and RoBERTa~\cite{liu2019roberta}) and thereby suffer from weak expressivity and interactive capabilities \cite{he2023survey}. Consequently, they fell short in adequately addressing complex clinical tasks, due to the limitations characterized by explainability, robustness, and generalization. The evolution of LLMs has dramatically reshaped this dilemma by inducing innovative capabilities that better align with the rigorous requirements of the clinical environment. For example, the emergent abilities of LLMs \cite{wei2022emergent} spark the few-shot and even zero-shot generalization capabilities, which significantly bolsters model explainability and thus address a critical concern in medical decision-making, \textit{e.g.}, Chain-of-Thought (CoT) methodology \cite{wei2022chain}. In essence, the progression from PLMs to LLMs signifies a leap forward in medical AI, which closes the gap between model capabilities and clinical environments. 

\begin{figure}[t]
\centering
\includegraphics[width=1\linewidth]{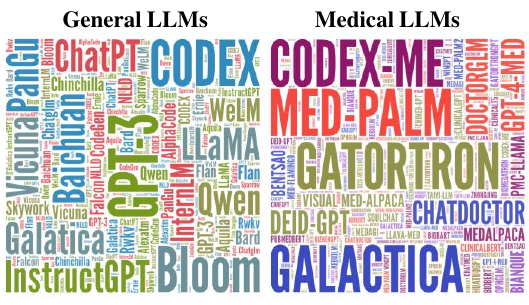}\\
\caption{Comparison of General LLMs and Medical LLMs. This chart highlights the diversity and specialization for large language models.}
\label{figure:model}
\end{figure}

Existing explorations of Med-LLMs mainly focus on identifying patient-specific factors to provide effective clinical decision-making support and treatment suggestions. These studies can be divided into the following key areas, where each one contributes uniquely to the broader understanding and advancement of Med-LLMs in healthcare.

$\bullet$ \textit{Medical Corpus:} Large-scale, high-quality medical corpus \cite{liu2024benchmarking} is vital for Med-LLMs to enhance their understanding of the medical terminology, context, and nuances of clinical language. Given the centrality of high-quality datasets, the direct idea is to collect more high-quality medical data from clinical scenarios, where a wide range of material resources are considered such as research papers, clinical case reports, medical textbooks, clinical guidelines, patient records, and drug information \cite{cai2024medbench}. The datasets should ensure that models are exposed to a diverse and representative spectrum of medical tasks. During data collection, some important challenges may arise from data privacy, standardization, and representation.

$\bullet$ \textit{Medical-specific Algorithm:} This branch focuses on the optimization of underlying learning paradigm from general purpose to specific domain. The relevant studies involve a series of adaptations and improvements to improve the LLMs' capacity for understanding medical language and context awareness. This includes domain-specific pre-training and fine-tuning to ensure that Med-LLMs could remain up-to-date with the latest medical knowledge and guidelines. Besides, for more accurate decision-making, some efforts are conducted to leverage structured medical knowledge graphs (\textit{e.g.}, KG-Rank \cite{yang2024kg}) to enhance Med-LLMs' clinical reasoning ability (\textit{e.g.}, Chain-of-Thought \cite{wei2022chain}), and to generate more reliable suggestions (\textit{e.g.}, retrieval-augmented generation).

\definecolor{paired-light-blue}{RGB}{198, 219, 239}
\definecolor{paired-dark-blue}{RGB}{49, 130, 188}
\definecolor{paired-light-orange}{RGB}{251, 208, 162}
\definecolor{paired-dark-orange}{RGB}{230, 85, 12}
\definecolor{paired-light-green}{RGB}{199, 233, 193}
\definecolor{paired-dark-green}{RGB}{49, 163, 83}
\definecolor{paired-light-purple}{RGB}{218, 218, 235}
\definecolor{paired-dark-purple}{RGB}{117, 107, 176}
\definecolor{paired-light-gray}{RGB}{217, 217, 217}
\definecolor{paired-dark-gray}{RGB}{99, 99, 99}
\definecolor{paired-light-pink}{RGB}{222, 158, 214}
\definecolor{paired-dark-pink}{RGB}{123, 65, 115}
\definecolor{paired-light-red}{RGB}{231, 150, 156}
\definecolor{paired-dark-red}{RGB}{131, 60, 56}
\definecolor{paired-light-yellow}{RGB}{231, 204, 149}
\definecolor{paired-dark-yellow}{RGB}{141, 109, 49}
\definecolor{paired-light-xxx}{RGB}{238 213 210}
\definecolor{paired-dark-xxx}{RGB}{139 123 139}
\definecolor{paired-light-yyy}{RGB}{175 238 238}
\definecolor{paired-dark-yyy}{RGB}{	176 196 222}
\tikzset{%
    parent/.style =          {align=center,text width=1.8cm,rounded corners=3pt, line width=0.3mm, fill=gray!10,draw=gray!80},
    child/.style =           {align=center,text width=2.3cm,rounded corners=3pt, fill=blue!10,draw=blue!80,line width=0.3mm},
    grandchild/.style =      {align=center,text width=2cm,rounded corners=3pt},
    greatgrandchild/.style = {align=center,text width=1.5cm,rounded corners=3pt},
    greatgrandchild2/.style = {align=center,text width=1.5cm,rounded corners=3pt},    
    referenceblock/.style =  {align=center,text width=1.5cm,rounded corners=2pt},
    top_class/.style =       {align=center,text width=2cm,rounded corners=3pt, fill=paired-light-gray!50,draw=paired-dark-gray!65,line width=0.3mm},
    top_class_1/.style =     {align=center,text width=2cm,rounded corners=3pt, fill=paired-light-gray!50,draw=paired-dark-gray!65,line width=0.3mm},
    tech/.style =            {align=center,text width=2cm,rounded corners=3pt, fill= paired-light-green!50,draw=paired-dark-green!75,line width=0.3mm}, 
    tech_more/.style =       {align=center,text width=3cm,rounded corners=3pt, fill= paired-light-green!50,draw=paired-dark-green!75,line width=0.3mm},   
    tech_work/.style =       {align=center,text width=5.0cm,rounded corners=3pt, fill= paired-light-green!50,draw= cyan!0,line width=0.3mm},
    develop/.style =         {align=center,text width=2cm,rounded corners=3pt, fill=paired-light-orange!50,draw=paired-dark-orange!65,line width=0.3mm},  
    develop_more/.style =    {align=center,text width=3cm,rounded corners=3pt, fill=paired-light-orange!50,draw=paired-dark-orange!65,line width=0.3mm}, 
    develop_work/.style =    {align=center,text width=5.0cm,rounded corners=3pt, fill=paired-light-orange!50,draw=red!0,line width=0.3mm},    
    task/.style =            {align=center,text width=2.0cm,rounded corners=3pt, fill=paired-light-pink!50,draw=paired-dark-pink!65,line width=0.3mm},  
    task_wide/.style =       {align=center,text width=3cm,rounded corners=3pt, fill=paired-light-pink!50,draw=paired-dark-pink!65,line width=0.3mm}, 
    task_work/.style =       {align=center,text width=5.0cm,rounded corners=3pt, fill=paired-light-pink!50,draw=paired-dark-pink!0,line width=0.3mm},    
    eval/.style =            {align=center,text width=2.0cm,rounded corners=3pt, fill=paired-light-yellow!50,draw=paired-dark-yellow!65,line width=0.3mm},  
    eval_wide/.style =       {align=center,text width=3cm,rounded corners=3pt, fill=paired-light-yellow!50,draw=paired-dark-yellow!65,line width=0.3mm}, 
    eval_work/.style =       {align=center,text width=5.0cm,rounded corners=3pt, fill=paired-light-yellow!50,draw=paired-dark-yellow!0,line width=0.3mm}, 
    specific/.style =        {align=center,text width=2.0cm,rounded corners=3pt, fill=paired-light-blue!50,draw=paired-dark-blue!65,line width=0.3mm},  
    specific_wide/.style =   {align=center,text width=8.5cm,rounded corners=3pt, fill=paired-light-blue!50,draw=paired-dark-blue!65,line width=0.3mm}, 
    challenge/.style =       {align=center,text width=2cm,rounded corners=3pt, fill=paired-light-blue!50,draw=paired-dark-blue!65,line width=0.3mm},
    challenge_wide/.style =  {align=center,text width=8cm,rounded corners=3pt, fill=paired-light-blue!50,draw=paired-dark-blue!65,line width=0.3mm},   
    algor/.style =           {align=center,text width=2.0 cm,rounded corners=3pt, fill= paired-light-xxx!50,draw=paired-dark-xxx!75,line width=0.3mm},   
    algor_wide/.style =           {align=center,text width=8.5cm,rounded corners=3pt, fill= paired-light-xxx!50,draw=paired-dark-xxx!75,line width=0.3mm}, 
    apply/.style =           {align=center,text width=2.0 cm,rounded corners=3pt, fill= paired-light-yyy!50,draw=paired-dark-yyy!75,line width=0.3mm},   
    apply_wide/.style =           {align=center,text width=8.5cm,rounded corners=3pt, fill= paired-light-yyy!50,draw=paired-dark-yyy!75,line width=0.3mm}, 
    trust/.style =           {align=center,text width=3cm,rounded corners=3pt, fill= paired-light-red!35,draw=paired-light-red!90,line width=0.3mm},       
}

\begin{figure*}[!t]%
\scriptsize
\centering
\begin{adjustbox}{center}
    \begin{forest}
        for tree={
            forked edges,
            grow'=0,
            draw,
            rounded corners,
            node options={align=center,},
            text width=3cm,
            s sep=9pt,
            calign=edge midpoint,
        },
        [\textbf{Medical LLMs}, fill=gray!45, parent
            [ Background \\ \S\ \ref{sec:background}, for tree={ top_class}
                [Development, for tree={fill=red!45,develop}
                    [Early NLP (Pre-2000s),  develop_more
                        [{Turing Test, RNNs, LSTM}, develop_work]
                    ]
                    [Deep Learning Revolution, develop_more 
                        [{Backpropagation for RNNs, Word2Vec by Mikolov et al., Transformer by Vaswani et al.}, develop_work]
                    ]
                    [Explosion of LLMs, develop_more 
                        [{GPT-3, GPT-4, LLaMA, ChatGPT, Bloom, Qwen, Baichuan, CodeX}, develop_work]
                    ]
                ]
                [Technology, for tree={fill=green!45, tech}
                    [Pre-Training, tech_more
                        [{NWD, MLM, RTD, NSP, SOP}, tech_work ]
                    ]
                    [Fine-tuning, tech_more
                        [{SFT, IFT, IPT, PEFT}, tech_work]
                    ]  
                    [{RLHF, ICL}, tech_more]  
                ]
            ]
            [ General2Specific \\ \S\ \ref{sec:General2Specific}, for tree={ top_class}
                 [Medical Task\&Data , for tree={fill=pink!45,task}
                     [Med-IE, task_wide
                         [{BC5CDR \cite{li2016biocreative}, CADEC \cite{karimi2015cadec}, NCBI \cite{dougan2014ncbi}, etc}, task_work ]
                     ]
                     [Med-QA, task_wide
                         [{MedQA \cite{jin2021disease}, PubMedQA \cite{jin2019pubmedqa}, MedMCQA \cite{pmlr-v174-pal22a}, etc}, task_work]
                     ]
                     [Med-NLI, task_wide
                         [{MedNLI \cite{romanov2018lessons}, etc}, task_work]
                     ]
                      [Med-Gen, task_wide
                         [{MIMIC-CXR \cite{johnson2019mimic}, PubMed \cite{sen2008collective}, MultiCochrane \cite{joseph2023multilingual}}, task_work]
                     ]
                 ] 
                 [Medical Evaluation, for tree={fill=pink!45,eval}
                    [Quantitative, eval_wide
                        [{BLEU, METEOR, ROUGE, CIDEr, Perplexity, etc}, eval_work]
                    ]
                    [Qualitative, eval_wide
                        [{Human Evaluation, Case Studies, User Feedback, etc.}, eval_work]
                    ]
                    [Automatic Evaluation, eval_wide
                        [{AutoEval \cite{liao2023automatic}, LLM-Mini-CEX \cite{shi2023llm}, MedGPTEval \cite{xu2023medgpteval}, RJUA-SPs \cite{liu2024towards}, etc}, eval_work]
                    ]
                ]
                [Specific Med-LLMs, for tree={fill=pink!45,specific}
                    [{ClinicalT5 \cite{lu_clinicalt5_2022}, ClinicalGPT \cite{wang2023clinicalgpt}, ChiMed-GPT \cite{tian2023chimed}, BioGPT \cite{luo2022biogpt}, PubMedBERT \cite{gu2021domain}, GatorTron~\cite{yang2022large}, Med-PaLM~\cite{singhal2023large}, MedAlpaca~\cite{han2023medalpaca}, LLaVA-Med~\cite{li_llava-med_2023}}, specific_wide
                    ]
                ]
            ]
            [Algorithm \\ \S\ \ref{sec:improve}, for tree={ top_class}
                [Clinical Reasoning, for tree={fill=yellow!45,algor}
                    [{ICP \cite{wu2024guiding}, JMLR \cite{wang2024jmlr}}, algor_wide]   
                ] 
                [Medical KG, for tree={fill=yellow!45,algor}
                    [{DR.KNOWS \cite{gao2023leveraging}, KG-Rank \cite{yang2024kg}, MedKgConv \cite{varshney2023knowledge}, ChiMed \cite{ye2023qilin}, DISC-MedLLM \cite{bao2023disc}},algor_wide] 
                ]
                [Medical Agent, for tree={fill=yellow!45,algor}
                    [{CT-Agent \cite{yue2024ct}, AutoGen \cite{wu2023autogen}, ArgMed-Agents \cite{hong2024argmed}, MAD \cite{smit2023we}},algor_wide] 
                ]
                [RAG, for tree={fill=yellow!45,algor}
                    [{Clinfo.ai \cite{lozano2023clinfo}, Almanac \cite{zakka2024almanac}, BiomedRAG \cite{li2024biomedrag}, Self-BioRAG \cite{jeong2024improving}, ECG-RAG \cite{yu2023zero}, ChatENT \cite{long2023chatent}, MIRAGE \cite{xiong2024benchmarking}, MedicineQA \cite{huang2024tool}},algor_wide] 
                ]
                [Human Alignment, for tree={fill=yellow!45,algor}
                    [{Safety Alignment \cite{han2024towards}, SELF-ALIGN \cite{sun2024principle}, EGR \cite{manathunga2023aligning}},algor_wide] 
                ]
                [Multi-Modal, for tree={fill=yellow!45,algor}
                    [{AD-MM-LLM \cite{feng2023large}, RAMM \cite{yuan2023ramm}, LLaVA-Med \cite{li_llava-med_2023}, Qilin-Med-VL \cite{liu2023qilin}},algor_wide] 
                ]
            ]
            [Application \\ \S\ \ref{sec:apply}, for tree={ top_class}
                [Applications, for tree={fill=yellow!45,apply}
                    [{Medical Diagnosis, Clinical Report Generation, Medical Education, Medical Robotics, Medical Language Translation}, apply_wide]   
                ] 
                [Challenges , for tree={fill=yellow!45,apply}
                    [{Protected Health Information, Clinical Workflows, Safety and Accountability},apply_wide] 
                ] 
            ]
            [Trustworthiness\\ \&Safety \\ \S\ \ref{sec:apply}, for tree={ top_class}
                [Fairness~\cite{li2023survey}, trust 
                ]
                [Accountability~\cite{solomon2023chatgpt}, trust 
                ]
                [Privacy~\cite{das2024security}, trust 
                ]
                [Robustness~\cite{alberts2023large}, trust 
                ]
            ]
            [Future Direction \\ \S\ \ref{future}, for tree={ top_class}
                [{Interpretability, Supportive policy, Clinical workflows, etc}, trust 
                ]
            ]
        ]      
    \end{forest}
\end{adjustbox}
\caption{Organization of the Survey on Medical Large Language Models. his detailed chart outlines the survey's structure, covering background, technology, medical tasks and data, evaluation methods, specific medical LLMs, algorithms, applications, trustworthiness and safety, and future directions. It traces the evolution from early NLP to the latest advancements in medical LLMs, including their development, pre-training, fine-tuning, and various evaluation metrics.}
\label{fig:paper_structure}
\end{figure*}

$\bullet$ \textit{Clinical Role:} Researchers also focus on the practical applications of Med-LLMs across various settings \cite{wu2023medical,wang2024beyond}, evaluating their impact on clinical workflows, patient outcomes, and healthcare efficiency. This includes assessing Med-LLMs in roles such as diagnostic support, treatment recommendation, patient communication, and medical education, and evaluating their effectiveness against traditional methods.

$\bullet$ \textit{Ethics, Privacy, and Interpretability:} With the increasing attention of medical AI, there is a concurrent need to address ethical concerns, ensure patient privacy, and establish regulatory guidelines \cite{mesko2023imperative,nori2023capabilities,yan2024protecting}. Research for this topic focuses on developing ethical Med-LLMs frameworks, safeguarding personal health information, and complying with legal requirements, such as GDPR \cite{data2016general} and HIPAA \cite{act1996health}. Besides, to enhance trustworthiness in AI-assisted decision-making, the explainability of Med-LLMs' outputs should be further enhanced to provide clear rationales for model predictions, \textit{e.g.}, enabling healthcare professionals to understand and evaluate the model's suggestions \cite{chu2024causal}.

Overall, through these advanced explorations in the field of Med-LLMs, various effective perspectives are delivered to promote the rapid development of medical AI societies. From the technical point, it is still a basic manner to improve the performance of Med-LLMs via increased model scale and complexity, which is consistent with the LLM's scaling law \cite{kaplan2020scaling}. The Med-LLMs, often pre-trained and fine-tuned on massive biomedical literature corpora like PubMed \cite{PubMedCentralTextMining}, outperform general-purpose language models on domain-specific tasks. To adapt to diverse clinical tasks, fine-tuning Med-LLMs over an array of specialized medical tasks is required.   
Essentially, to track the potential pathways for developing Med-LLMs, several reviews \cite{thirunavukarasu2023large,omiye2024large} primarily focus on exploring the clinical capabilities and applications of Med-LLMs, including Electronic Health Records (EHRs) \cite{yang2022gatortron}, health education \cite{kung2023performance}, and diagnostic decision making \cite{kottlors2023feasibility}.

$\triangle$ The study \cite{he2023survey} outlines the healthcare capabilities of LLMs, providing a developing roadmap from traditional Pre-trained Language Models (PLMs) to LLMs. 

$\triangle$ The study \cite{thirunavukarasu2023large} summarizes how LLMs are developed (\textit{e.g.}, ChatGPT) and being leveraged in clinical settings. 

$\triangle$ The study \cite{huang2024comprehensive} targets to provide an in-depth analysis of LLM applications in the medical industry across clinical tasks, research, and education.

Nevertheless, despite the promising progress and detailed reviews, the potential pathways of Med-LLMs are still under-explored, \textit{e.g.}, aligning the development of Med-LLMs with the complex needs in the clinical environment, which is vital for better patient care and advancing medical research. Firstly, compared with the latest studies about Med-LLMs, a more well-targeted literature review is still needed to delve into both technical (medical task, data, evaluation, and algorithm) and social impacts (application, trustworthiness, and safety). Secondly, due to the explosive growth of Med-LLMs, it is nontrivial for the research community to conduct a both comprehensive and fine-grained exploration of existing Med-LLMs, covering historical background and technology innovations. Thirdly, it is challenging to align Med-LLMs with human (or clinicians) preferences or values, due to the risk of producing toxic, fictitious, or harmful content. It should consider trustworthy and ethical constraints to ensure fairness, accountability, privacy, and robustness in Med-LLMs applications. All these aspects are pivotal for the responsible development of Med-LLMs, which can effectively address clinical needs in practice with enhanced patient care, ethical boundaries, and trustworthiness.

To identify the most pressing needs and challenges faced by healthcare professionals, patients, and researchers, this survey aims to conduct a comprehensive literature review of the latest advances in Med-LLMs from various perspectives. We thoroughly collect the literature and summarize the background, basic techniques, data, algorithm, application, and challenges for Med-LLMs. As shown in Figure \ref{fig:paper_structure}, this review seeks to provide the following explorations and investigations. \textbf{Section 2:} The historical background and fundamental training techniques. \textbf{Section 3:} From general-purpose to medical-specific LLMs. \textbf{Section 4:} Improving algorithms for Med-LLMs. \textbf{Section 5:} Feasible applications of Med-LLMs. \textbf{Section 6:} Discussion about trustworthy and safety considerations.


\begin{figure*}[t]
\centering
\includegraphics[width=0.9\linewidth]{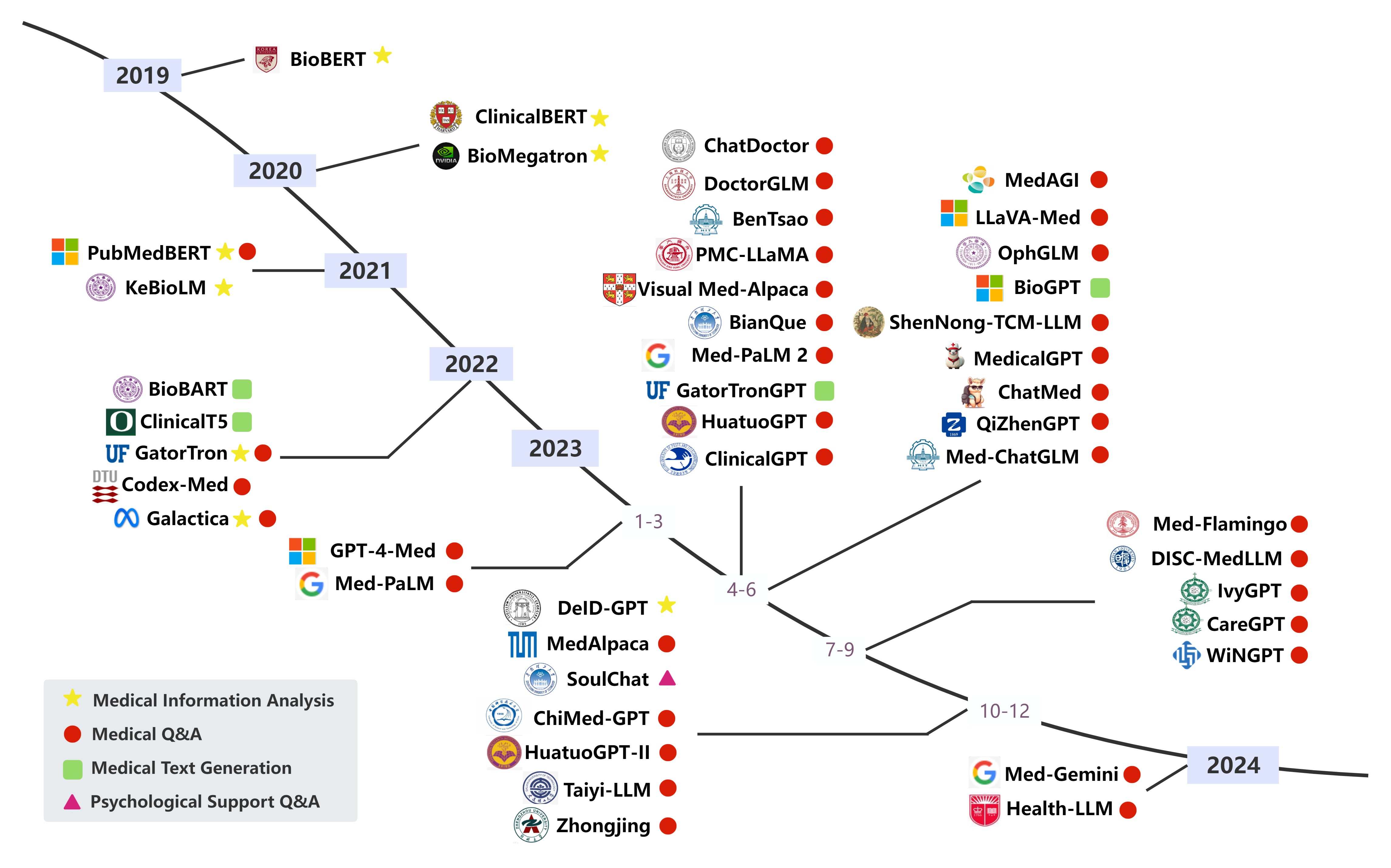}\\
\caption{Evolution of Medical Large Language Models from 2019 to 2024. This roadmap highlights the progression and diversification of medical large language models over the years. Starting with BioBERT in 2019, the field has seen the emergence of specialized models for medical information analysis, Q\&A, text generation, and psychological support. Notable developments include ClinicalBERT, BioMegatron, and PubMedBERT in 2020, followed by a surge in models like ChatDoctor, DoctorGLM, and Visual Med-Alpaca in 2022. The roadmap culminates with advanced models such as Med-Gemini and Health-LLM in 2024, reflecting the ongoing innovation in leveraging AI for healthcare applications.}
\label{figure:timeline}
\end{figure*}

\section{Background and Technology}
\label{sec:background}

\subsection{Model Architecture}
The Transformer \cite{vaswani2017attention} is a groundbreaking architecture primarily designed for sequence-to-sequence tasks, such as machine translation \cite{wu2024adapting}. The transformer-based models have grown exponentially in size over the years, with billions to hundreds of billions of parameters. The core components of the transformer include the following aspects.

\subsubsection{Encoder-Decoder Structure} The Transformer consists of two main parts, an encoder that processes the input sequence, and a decoder that generates the output sequence. This separation allows for flexible input-output handling typical in tasks like translation. Each layer in the encoder (or decoder) typically consists of two main parts: a Multi-Head Self-Attention Mechanism followed by a Position-wise Feed-Forward Network (FFN). This structure is repeated multiple times (typically 6 layers in the original Transformer paper, but modern models may use dozens or even hundreds of layers) to form a stack within both the Encoder and Decoder.

\subsubsection{Self-Attention Mechanism} A self-attention layer computes attention scores for each input token with respect to other tokens in the sequence. To capture diverse relationships within the data, this process is performed in parallel multiple times with different linear projections for queries, keys, and values, allowing the model to attend to different aspects of the input concurrently, named multi-head self-attention. Each head focuses on different parts of the input, enhancing the model's ability to grasp intricate structures.

\subsubsection{Positional Encoding} Positional Encoding is an important component in the transformer for incorporating the order information of the sequence since the operations in the transformer lack the ability to provide position information in sequential data. The commonly used positional encoding strategies include sinusoidal conditional encoding, learned positional embeddings, relative positional encoding, rotary conditional encoding, and hybrid positional embedding.

\subsubsection{Residual Connections \& Layer Normalization} For each layer, residual connections are employed to mitigate the problem of vanishing gradients during backpropagation. After the residual connection, layer normalization is applied to stabilize and speed up the training process. This structure ensures that the model can effectively learn from very deep architectures without suffering from degradation issues.

\subsection{Training Techniques}
\subsubsection{Discussion on Pre-training Stage} LLM pre-training refers to the process where a model is initially trained on a large and general-purpose corpus of text data before being fine-tuned for specific tasks. The pre-training tasks are important for learning the universal and general representation of language, which basically contribute to significant improvements in model performance via a self-supervised manner. Here, we summarize some commonly used pre-training tasks in Table \ref{tab:pretrain-survey}, where the detailed introductions are as follows:

$\bullet$ \textit{Next Word Prediction (NWP)} \cite{howard2018universal}. Known as Language Modeling, NWP aims to predict the most possible next word in a sequence given the previous words. At its core, NWP relies on understanding the context provided by the preceding words. This context helps the model infer the statistical likelihood of different words following in the sequence, \textit{i.e.}, given a huge corpus, the entire LLM is optimized with maximum likelihood estimation (MLE).

$\bullet$ \textit{Mask Language Modeling (MLM)} \cite{devlin2018bert}. MLM involves training a model to predict the original identities of certain words (tokens) in a sentence that have been randomly replaced with a special mask token (often ``[MASK]"). By predicting the missing words based on remained context, the model can learn to understand the semantic and bidirectional relationships between words. The basic masking strategy involves masking 15\% of the tokens in a sequence. Not all masked tokens are treated equally. For the masked tokens, 80\% are replaced by the ``[MASK]" token, 10\% are replaced by a random word, and 10\% are left unchanged. This variation prevents the model from simply memorizing the masked positions and forces it to learn more generalizable context representations.

\begin{table*}[t]
 \centering
 \begin{threeparttable}
 \caption{Illustration for Different Pre-training Tasks (from \cite{qiu2020pre}). $\mathbf{x}=[x_1,x_2,\cdots,x_T]$ denotes a sequence.}
 \label{tab:pretrain-survey}
 \begin{tabular}{l|l|l}
 \hline
   \textbf{Pre-training Task} & \textbf{Objective Function} & \textbf{Description}\\%
   \hline %
   LM& $\displaystyle\mathcal{L}_{\textrm{\tiny LM}}=-\sum_{t=1}^{T} \log p(x_t|\mathbf{x}_{<t})$ & $\mathbf{x}_{<t} = x_1,x_2,\cdots,x_{t-1}$.\\
   \hline
   MLM& $\displaystyle \mathcal{L}_{\textrm{\tiny MLM}}=-\sum_{\hat{x}\in m(\mathbf{x})} \log p\Big(\hat{x}|\mathbf{x}_{\setminus m(\mathbf{x})}\Big)$ &$m(\mathbf{x})$ and $\mathbf{x}_{\setminus m(\mathbf{x})}$ denote the masked words from $\mathbf{x}$ and the rest words respectively.\\
   \hline
   NSP/SOP & $\displaystyle \mathcal{L}_{\textrm{\tiny NSP/SOP}}=-\log p(t|\mathbf{x},\mathbf{y})$ & $t=1$ if $\mathbf{x}$ and $\mathbf{y}$ are continuous segments from corpus.\\
   \hline
   RTD & $\displaystyle \mathcal{L}_{\textrm{\tiny RTD}}=-\sum_{t=1}^{T} \log p(y_t|\hat{\mathbf{x}})$ & $y_t=\mathbf{1}(\hat{x}_t=x_t)$, $\hat{\mathbf{x}}$ is corrupted from $\mathbf{x}$.\\
 \hline
 \end{tabular}
 \end{threeparttable}
\end{table*}

$\bullet$ \textit{Replaced Token Detection (RTD)} \cite{clark2020electra}. RTD is a pre-training methodology designed to enhance language models' capability to discern subtle semantic variations in the text by detecting tokens that have been deliberately substituted with synonyms or semantically equivalent terms. This approach leverages the richness of linguistic synonyms and near-synonyms to create a learning environment that sharpens the model's sensitivity to the nuances of meaning. During training, two types of sentence instances are generated: (1) Unaltered Sentences: A portion of sentences remain unmodified, serving as a baseline for the model to learn the standard usage of words in context. (2) Altered Sentences: The remaining sentences undergo a careful replacement of selected words with synonyms, creating alternative versions that maintain the overall message but differ in lexical choice. By integrating RTD into the pre-training regimen, models can acquire a more nuanced understanding of semantic similarity and distinction. This refined understanding proves beneficial for downstream tasks that demand high levels of semantic precision, such as text entailment, sentiment analysis, and semantic textual similarity tasks, where the ability to differentiate between subtle shifts in meaning is crucial.

$\bullet$ \textit{Next Sentence Prediction (NSP)}  \cite{devlin2018bert}. Punctuation naturally segments text, offering a logical foundation for developing pre-training strategies based on this inherent structure. The NSP task is designed to train models to discern whether a pair of sentences presented to them are contiguous segments taken from a text corpus. In practice, during NSP's pre-training phase, two kinds of pairs are constructed:
(1) Continuous Pairs (IsNext): Approximately 50\% of the time, the model is given sentence pairs that are genuinely consecutive in the original text.
(2) Random Pairs (NotNext): For the remaining 50\%, the second sentence in the pair is randomly selected from elsewhere in the corpus, forming unrelated or discontinuous sequences. This balanced approach ensures that models not only learn to recognize sequential sentence structures but also develop an understanding of contextual continuity versus disjointed pairings.

$\bullet$ \textit{Sentence Order Prediction (SOP)} \cite{lan2019albert}. As a pre-training task, SOP can improve a model's understanding of the coherence and narrative flow in a text by determining the correct sequential order of sentences, which is particularly useful in training models for tasks that require comprehension of long-form text, dialogue understanding, or any scenario where the sequence of events matters. The primary goal of SOP is to teach models to recognize the logical progression of ideas within a text by predicting whether a pair of sentences are arranged in their original, sequential order or not. Thus, two types of sentence pairs are formulated:
(1) Sequential Pairs: Half of the time, sentence pairs are kept in their original order as they appear in the text. These form the positive examples, labeled as `IsOrder'. (2) Non-Sequential Pairs: The remaining half are constructed by swapping the order of sentences from different parts of the text, creating incoherent or unrelated sequences. These are labeled as `NotOrder', serving as negative examples. The model is fed these sentence pairs and tasked with predicting whether the given order is correct according to the narrative flow. In this way, SOP encourages models to learn the underlying logic and coherence that connects sentences, enabling them to infer the correct sequence of events or thoughts.

\begin{table*}[!t]
\caption{Basic Medical NLP Tasks. These tasks often involve analyzing, understanding, and extracting meaningful insights from clinical notes, medical records, research papers, and patient reports.}
\label{basic_task}
\centering
\begin{tabular}{c|m{4cm}|m{9cm}}
\hline
\textbf{Medical Task}            &\textbf{Sub-task}                     & \textbf{Description}  \\ \hline 
\multirow{5}{*}{\textbf{Med-IE}} &Entity Recognition \cite{li2020survey}& Identifying medical concepts such as diseases, symptoms, and treatments \cite{abacha2011medical}.\\ \cline{2-3}
                                 &Relationship Extraction \cite{zhou2005exploring} & Detecting relationships between entities \cite{goel2023llms}.\\ \cline{2-3}
                                 &Event Extraction \cite{liu2020event} & Recognizing clinical events and attributes, \textit{e.g.}, onset and duration \cite{yang2020clinical}.\\ \cline{2-3}
                                 &Information Summarization \cite{kim2023sure} & Condensing large medical records into concise and critical information \cite{tang2023evaluating}.\\ \cline{2-3}
                                 &Adverse Drug Event Detection \cite{fan2020adverse}& Identifying potential adverse reactions or side effects of medications \cite{webster2023six}.\\ \hline
\multirow{7}{*}{\textbf{Med-QA}} &Query Understanding \cite{wang2023large-a}& find out the meaning of the question proposed by users. \\ \cline{2-3}
                                 &Information Retrieval \cite{zhuang2024toolqa} & searching through databases to retrieve relevant information.\\ \cline{2-3}
                                 &Inference and Reasoning \cite{chen2023evaluation} & Based on the extracted information, the model may need to perform reasoning tasks, such as inferring relationships between medical concepts, determining causality, or predicting outcomes. \\ \hline
\multirow{5}{*}{\textbf{Med-NLI}} & Textual Entailment\cite{tawfik_towards_2019} & Determining whether the hypothesis logically follows from the premise.\cite{yadav_medical_2020} \\ \cline{2-3}
& Contradiction Detection\cite{yazi_experimental_2021} & Identifying when the hypothesis contradicts the information in the premise. \\ \cline{2-3}
& Neutral Relationship Identification & Recognizing when the premise and hypothesis are not semantically related enough to entail or contradict each other. \\ \cline{2-3}
& Causality Recognition\cite{yang_survey_2022} &Inferring causal relationships between events in the premise and hypothesis.\\ \hline
\multirow{1}{*}{\textbf{Med-Gen}} & Content Generation\cite{zhou_evaluation_2023} & generating new medical descriptions or knowledge based on a given input\\ \hline
\end{tabular}
\end{table*}

\subsubsection{Discussion on Fine-tuning Stage}
Fine-tuning a LLM involves adapting a pre-trained model to a specific task or domain. This process can help improve the model's performance on tasks using a smaller and specialized dataset. The commonly used fine-tuning techniques are as follows:

$\bullet$ \textit{Supervised Fine-tuning (SFT)}~\cite{taori2023stanford, lu2023instag}. SFT follows a structured approach to adapt pre-trained models to specific tasks by further training them on new, labeled datasets. In SFT, the pre-trained LLM is fine-tuned on labeled datasets using supervised learning techniques.

$\bullet$ \textit{Instruction Fine-tuning (IFT)} \cite{zhang2023instruction}. IFT, also known as prompt-based fine-tuning or instruction tuning, is a recent adaptation of the traditional fine-tuning method, particularly popularized in the context of LLMs. It focuses on teaching models to follow instructions or prompts to perform a variety of tasks without explicit task-specific architecture changes or separate training phases for each task. Instead of fine-tuning on a single-task dataset, instruction fine-tuning involves using a diverse set of instructions or prompts that cover multiple tasks within the same training process. Besides, a medical IFT variant is proposed called Instruction Prompt Tuning (IPT) to train Med-PaLM.

$\bullet$ \textit{Parameter-Efficient Fine-ting (PEFT)} \cite{han2024parameter}. PEFT refers to techniques designed to adapt large pre-trained models to specific downstream tasks while minimizing the introduction of new parameters or modifying only a fraction of the existing ones. This approach aims to retain most of the pre-training knowledge, reduce computational costs, and alleviate the risk of overfitting, especially when the target task dataset is small. One popular and classical PEFT strategy is LoRA \cite{hu2021lora}, which introduces low-rank matrices that are multiplied with the weight matrices of specific layers during forward and backward passes. This allows the model to learn task-specific adjustments without changing the original weights directly, reducing the number of new parameters needed.

\subsubsection{Reinforcement Learning from Human Feedback} RLHF techniques ~\cite{ouyang2022training, touvron2023llama} focus on collecting human feedback on the output of model generation to guide the model for further optimization. This approach aims to train AI agents to perform tasks more aligned with human desires and ethical standards, especially in complex, real-world environments where manually designing reward functions is challenging or insufficient. 

\subsubsection{In-Context Learning}. Also known as few-shot learning for LLMs, ICL \cite{dong2022survey} refers to the capability of generalizing and adapting to new tasks by providing examples within the prompt, without requiring additional fine-tuning on a specific training dataset. This method leverages the model's pre-existing knowledge and its ability to understand the context to infer what is being asked and generate appropriate responses.

\begin{table*}[!t]
\caption{Statistics Information and Resources of Medical NLP Dataset. The data scale is the number of sentences or Q-A pairs. The year refers to the publication date.}
\label{llm_data}
\centering
\begin{tabular}{c|l|c|c|c|c|c}
\hline
\textbf{Medical Task} & \textbf{Dataset}  &\textbf{Year}  &\textbf{Scale}          &\textbf{Language}            &\textbf{Task}              &\textbf{Link}\\ \hline 
\multirow{10}{*}{\textbf{Med-IE}} 
&GENIA \cite{kim2003genia}                &2003  &$\sim$18.5K     &English         &NER   &\href{https://huggingface.co/datasets/Rosenberg/genia}{Link}\\ 
&GENIA11 \cite{kim2011overview}           &2011  &$\sim$10K       &English         &MEE   &\href{https://2011.bionlp-st.org/}{Link}\\ 
&ADE \cite{gurulingappa2012development}   &2012  &$\sim$4K        &English         &RE    &\href{https://groups.csail.mit.edu/vision/datasets/ADE20K/}{Link}\\ 
&ShARe13 \cite{pradhan2013task}           &2013  &$\sim$29K       &English         &NER   &\href{https://healthnlp.hms.harvard.edu/share/wiki/index.php/Main_Page}{Link}\\ 
&GENIA13 \cite{kim2013genia}              &2013 &$\sim$5K         &English         &EE   &\href{https://bmcbioinformatics.biomedcentral.com/articles/supplements/volume-16-supplement-10}{Link}\\
&NCBI \cite{dougan2014ncbi}               &2014  &$\sim$7K        &English         &NER   &\href{https://www.ncbi.nlm.nih.gov/research/bionlp/Data/disease/}{Link} \\ 
&ShARe14 \cite{mowery2014task}            &2014  &$\sim$35K       &English         &NER   &\href{https://clefehealth.imag.fr/?page_id=455}{Link}\\ 
&CADEC \cite{karimi2015cadec}             &2015  &$\sim$7.5K      &English         &NER   &\href{https://data.csiro.au/collection/csiro:10948?v=3&d=true}{Link} \\ 
&BC5CDR \cite{li2016biocreative}          &2016  &$\sim$14K       &English         &NER   &\href{https://huggingface.co/datasets/bigbio/bc5cdr}{Link} \\ 
&PHEE \cite{sun2022phee}                  &2022  &$\sim$5K       &English           &EE   &\href{https://github.com/zhaoyuesun/phee}{Link}\\
\hline
\multirow{23}{*}{\textbf{Med-QA}}
&emrQA \cite{pampari2018emrqa}       &2018  &$\sim$1B       &English             &QA   &\href{https://emrqa.github.io/}{Link} \\ 
&Medical DS \cite{wei2018task} &2018  &-     &Chinese             &Dialogue   &\href{http://www.sdspeople.fudan.edu.cn/zywei/data/acl2018-mds.zip}{Link}\\
&MedicationQA \cite{abacha2019bridging} &2019   &$\sim$674  &English             &QA   &\href{https://github.com/abachaa/Medication_QA_MedInfo2019}{Link}\\
&MedQuAD \cite{BenAbacha-BMC-2019}   &2019  &$\sim$47K      &English             &QA   &\href{https://github.com/abachaa/MedQuAD}{Link}\\
&webMedQA \cite{he2019applying}      &2019  &$\sim$63K      &Chinese             &QA   &\href{https://github.com/hejunqing/webMedQA}{Link} \\ 
&PubMedQA \cite{jin2019pubmedqa}     &2019  &$\sim$280K     &English             &Multiple-choice   &\href{https://pubmedqa.github.io/}{Link}\\ 
&LiveQA \cite{liu2020liveqa}          &2020  &$\sim$117K     &Chinese             &Multiple-choice   &\href{https://github.com/PKU-TANGENT/LiveQA}{Link}\\
& MedDialog \cite{zeng2020coviddialogmodel}&2020 &$\sim$3.66M&Chinese \& English&Dialogue&\href{https://github.com/UCSD-AI4H/COVID-Dialogue}{Link}\\
& CovidDialog \cite{zeng2020coviddialogmodel}&2020&$\sim$600&Chinese \& English&Dialogue&\href{https://github.com/UCSD-AI4H/COVID-Dialogue}{Link}\\ 
& MEDIQA \cite{savery2020question}&2020&$\sim$2K&English&Dialogue&\href{https://huggingface.co/datasets/medalpaca/medical\_meadow\_wikidoc\_patient\_information}{Link}\\
& CORD-19 \cite{wang2020cord}&2020&$\sim$1M&English&Dialogue&\href{https://huggingface.co/datasets/medalpaca/medical\_meadow\_cord19}{Link}\\
&MMLU \cite{hendryckstest2021}       &2021  &$\sim$116K     &English             &Multiple-choice   &\href{https://github.com/hendrycks/test}{Link}\\ 
&MedQA \cite{jin2021disease}         &2021  &$\sim$270K     &Chinese \& English  &Multiple-choice   &\href{https://github.com/jind11/MedQA}{Link}\\
&CMCQA \cite{xia-etal-2022-medconqa} &2022  &$\sim$20M      &Chinese             &QA   &\href{https://github.com/WENGSYX/CMCQA}{Link}\\ 
&MedMCQA \cite{pmlr-v174-pal22a}     &2022  &$\sim$193K     &English             &Multiple-choice   &\href{https://medmcqa.github.io/}{Link}\\ 
&HealthSearchQA \cite{singhal2023large} &2022 &$\sim$3K     &English             &QA   &N/A\\
& ChatDoctor \cite{li2023chatdoctor} &2023&$\sim$200K&English&Dialogue&\href{https://github.com/Kent0n-Li/ChatDoctor}{Link}\\
&Huatuo-26M \cite{li2023huatuo}      &2023  &$\sim$26M      &Chinese             &QA   &\href{https://github.com/FreedomIntelligence/Huatuo-26M}{Link} \\ 
& Wikidoc Patient Information &2023&$\sim$6K&English&Dialogue&\href{https://huggingface.co/datasets/medalpaca/medical\_meadow\_wikidoc\_patient\_information}{Link}\\ 
& Medical Flashcards \cite{han2023medalpaca}&2023&$\sim$34K&English&Dialogue&\href{https://github.com/kbressem/medalpaca}{Link}\\ 
& Wikidoc&2023&$\sim$67K&English&Dialogue&\href{https://huggingface.co/datasets/medalpaca/medical\_meadow\_wikidoc}{Link}\\ 
&RJUA-QA \cite{BenAbacha-BMC-2019}   &2023  &$\sim$2K      &Chinese              &QA   &\href{https://github.com/alipay/RJU_Ant_QA}{Link}\\
\hline
\multirow{1}{*}{\textbf{Med-NLI}} 
&MedNLI \cite{romanov2018lessons}          &2018  &$\sim$14K       &English         &NLI   &\href{https://jgc128.github.io/mednli/}{Link} \\ \hline
\multirow{1}{*}{\textbf{Med-Gen}} 
&PubMed \cite{sen2008collective} & 2008 &19.7K articles   &English         &Text Summarization   &\href{https://physionet.org/content/mimiciii/1.4/}{Link} \\
&MIMIC-III \cite{johnson2016mimic} & 2016 &73K   &English         &Text Summarization   &\href{https://physionet.org/content/mimiciii/1.4/}{Link} \\
&MIMIC-CXR \cite{johnson2019mimic} & 2019 &128K   &English         &Text Summarization   &\href{https://physionet.org/content/mimic-cxr/2.0.0/}{Link} \\
&MeQSum \cite{abacha2019summarization} & 2019 &1K pairs   &English         &Text Summarization   &\href{https://github.com/abachaa/MeQSum}{Link} \\
&CORD-19 \cite{wang2020cord} & 2020 &140K articles  &English         &Text Summarization   &\href{https://github.com/allenai/cord19}{Link} \\
&MentSum \cite{sotudeh2022mentsum} & 2022 &24K pairs  &English         &Text Summarization   &\href{https://ir.cs.georgetown.edu/resources/}{Link} \\
&MultiCochrane \cite{joseph2023multilingual} & 2023 &7.8K pairs  &English &Text Simplification & \href{https://github.com/SebaJoe/MultiCochrane}{Link}   \\      
&PMC \cite{PubMedCentralTextMining} &  Update &9,407K articles   &English         &Text Summarization   &\href{https://www.ncbi.nlm.nih.gov/pmc/}{Link} \\
\hline
\end{tabular}
\end{table*}

\section{From General to Medical-Specific LLMs}
\label{sec:General2Specific}

\subsection{NLP Tasks under Medical Domain}
In this section, we summarize the primary LLMs' tasks in the medical domain, including Medical Information Extraction (Med-IE), Question-Answer (Med-QA), Natural Language Inference (Med-NLI), and Medical Text Generation (Med-Gen). These applications are fundamental NLP tasks but extremely challenging due to the complexity of the medical domain.

$\bullet$ \textit{Medical Information Extraction.}
Medical Information Extraction (Med-IE) is a fundamental field to extract relevant medical information from unstructured or semi-structured sources such as electronic health records (EHRs), clinical notes, medical research articles, and other healthcare documents. The aim is to transform this textual data into structured formats that can be easily analyzed, shared, and utilized in various applications like decision support systems, population health management, and medical research. The common applications of Med-IE are summarized in Table \ref{basic_task}.

When applied LLMs for the Med-IE task, the powerful generalization ability can well deal with more complex conditions and support more convenient applications. For instance, benefiting from the zero- and few-shot capabilities, InstructGPT \cite{ouyang2022training} can be helpful in performing information extraction from clinical text \cite{agrawal2022large}, such as biomedical evidence extraction \cite{nye2018corpus} and medical status extraction \cite{moon2014sense}, although InstructGPT is not trained specifically for the medical domain. 

$\bullet$ \textit{Medical Question Answer.}
The Medical Question Answering (Med-QA) task is a specific application for building AI systems (\textit{e.g.}, LLMs) capable of understanding and answering complex medical questions from patients, healthcare providers, or researchers. The system typically leverages large amounts of structured and unstructured data sources such as electronic health records, medical textbooks, research articles, and online resources to generate accurate and reliable answers. Here, we summary how the Med-QA system works in the Table \ref{basic_task}. Information retrieval could involve searching through medical literature, patient histories, or other knowledge bases, which aims to identify and extract key pieces of evidence that are pertinent to the question.

$\bullet$ \textit{Medical Natural Language Inference.}
Medical Natural Language Inference  (Med-NLI) is to deal with understanding the relationship between two pieces of medical text and determining whether one sentence logically entails, contradicts, or is neutral with respect to another. This task is also known as Recognizing Textual Entailment (RTE) in the broader context of NLP. In Med-NLI, the inputs are typically pairs of sentences, where one is called the premise and the other is the hypothesis. For example: 

\textit{Premise:} ``The patient has a history of hypertension." 

\textit{Hypothesis:} ``The patient suffers from high blood pressure." 

In this case, the inference system would determine that the hypothesis is entailed by the premise because hypertension is a medical term for high blood pressure. The importance of Med-NLI lies in its ability to help machines understand and reason about medical knowledge expressed in natural language. It can be used in various applications, such as Medical Information Validation, Decision Support Systems, and Medical Knowledge Base Construction and Maintenance.

$\bullet$ \textit{Medical Text Generation}
Medical Text Generation (Med-Gen), specifically focusing on tasks like text summarization, is an application of artificial intelligence where algorithms process and condense lengthy medical documents into concise summaries. This is particularly useful in the healthcare sector where professionals deal with vast amounts of complex information daily, including clinical research papers, patient records, diagnostic reports, and treatment guidelines. The primary goal is to save time and improve efficiency by extracting the most critical information from lengthy texts. Summaries can help doctors, researchers, and other healthcare professionals quickly understand the essence of a study, patient history, or clinical guideline without having to read through the entire document.

\begin{table*}[!t]
\centering
\caption{Overview of Medical Large Language Models (Med-LLMs). This table summarizes key Med-LLMs, detailing their release year, training methods, training data, and evaluation datasets used to assess their performance in various medical tasks.}
\label{tab: med-llms}
\resizebox{\linewidth}{!}{
\begin{tabular}{l|l|l|l|p{7cm}}
\hline
\textbf{Model}          & \textbf{Year}       & \textbf{Method}     & \textbf{Training Data}                       & \textbf{Evaluation Data}                                 \\ \hline
BioBERT~\cite{lee2020biobert}                    &2019 &PT      & Medical Abstracts and Articles &NER, RE , QA\\
ClinicalBERT~\cite{huang2019clinicalbert}        &2020 &PT      & Clinical notes &MIMIC-III \\
BioMegatron~\cite{shin2020biomegatron}           &2020 &PT      & Medical Literature &BC5CDR, NCBI, ChemProt, BioASQ-7b-factoid\\
PubMedBERT~\cite{gu2021domain}                   &2021 &PT      & Medical Literature &BLURB \\
KeBioLM~\cite{yuan_improving_2021}               &2021 &PT & Medical Literature  &BLURB\\
BioBART~\cite{yuan_biobart_2022}                 &2022 &PT & Medical Literature &CovidDialog, iCliniq, MeQSum, etc. \\
ClinicalT5~\cite{lu_clinicalt5_2022}             &2022 &PT, & Medical Records&UMLS, HOC, NCBI, BC5CDR, MEDNLI\\
GatorTron~\cite{yang2022gatortron}               & 2022 & PT         & Clinical Notes                      & CNER, MRE, MQA                                \\ 
Codex-Med~\cite{lievin2024can}                   & 2022 & ICL        &Programming Code& USMLE, MedMCQA,PubMedQA                       \\ 
Galactica~\cite{taylor2022galactica}             & 2022 & PT, IFT    & DNA Sequence                        & MedMCQA, PubMedQA, Medical Genetics           \\ 
Med-PaLM~\cite{singhal2023large}                 & 2022 & IPT         & Medical Datasets                        & MultiMedQA, HealthSearchQA                    \\ 
GPT-4-Med~\cite{nori2023capabilities}            & 2023 & ICL        &General Open-source Data& USML E, MultiMedQA                            \\ 
DeID-GPT~\cite{liu2023deid}                      & 2023 & ICL        &General Open-source Data& i2b2/UTHealth de-identification task        \\ 
ChatDoctor~\cite{yunxiang2023chatdoctor}         & 2023 & IFT        & Patient-doctor Dialogues            & iCliniq                                       \\ 
DoctorGLM~\cite{xiong2023doctorglm}              & 2023 & IFT        & Chinese medical Dialogues           & -                                             \\ 
MedAlpaca~\cite{han2023medalpaca}                & 2023 & IFT        & Medical Dialogues and QA            & USMLE, Medical Meadow                         \\ 
BenTsao~\cite{wang2023huatuo}                    & 2023 & IFT        & Medical KG, Medical QA & Customed medical QA                           \\ 
PMC-LLaMA~\cite{wu2023pmc}                       & 2023 & IFT        & Biomedical Academic Papers          & PubMedQA, MedMCQA, USMLE                      \\ 
Visual Med-Alpaca~\cite{shu2023visual}           & 2023 & PT, IFT    & Medical QA                          &-                                               \\ 
BianQue~\cite{chenbianque}                       & 2023 & IFT        & Medical QA &MedDialog-CN,MedDG,CHIPMDCFNPC,IMCS-V2\\ 
Med-PaLM 2~\cite{singhal2023towards}             & 2023 & IFT        &  MedMCQA, PubMedQA, etc.                                   & MultiMedQA, Long-form QA                      \\ 
GatorTronGPT~\cite{peng2023study}                & 2023 & PT         & Clinical and General Text           & PubMedQA, USMLE, MedMCQA, DDI, KD-DTI \\ 
HuatuoGPT~\cite{zhang2023huatuogpt}              & 2023 & IFT        & Instruction and Conversation Data   & CmedQA, webmedQA, and Huatuo26M               \\ 
ClinicalGPT~\cite{wang2023clinicalgpt}           & 2023 & IFT+RLHF   & Medical dialogues and QA, EHR       & MedDialog, MEDQA-MCMLE, MD-EHR       \\ 
MedAGI~\cite{zhou2023path}                       & 2023 & IFT        & Public Medical Datasets and Images  & SkinGPT-4, XrayChat, PathologyChat            \\ 
LLaVA-Med~\cite{li_llava-med_2023}               & 2023& IFT        & Multimodal Biomedical Instruction   & VQA-RAD, SLAKE, PathVQA                       \\ 
OphGLM~\cite{gao2023ophglm}                      & 2023 & IFT        & KG, Medical Dialogues & Fundus diagnosis pipeline tasks      \\ 
SoulChat~\cite{chen2023soulchat}                 & 2023 & IFT        & Long Text, Empathetic Dialogue      &SoulChatCorpus,SMILECHAT\\ 
Med-Flamingo~\cite{moor2023med}                  & 2023 & IFT        & Image caption/tokens pairs          & VQA-RAD, Path-VQA, Visual USMLE               \\ 
BioGPT~\cite{luo2022biogpt}                      & 2023 &PT, IFT & Medical Literature &BC5CDR, KDDTI, DDI, PubMedQA, HOC \\
ChiMed-GPT~\cite{tian2023chimed}                 & 2023 &PT, RLHF  & Medical Encyclopedia and Articles&CCKS-2019, ChiMST, C-Eval, CMMLU, MedQA \\
DISC-MedLLM~\cite{bao2023disc}                  &2023 &IFT  & Existing Medical Datasets &CMD, CMID, MLEC-QA, NEEP 306\\
IvyGPT~\cite{wang_ivygpt_2023}                  &2023 &RLHF & Instruction and Conversation Data & 100 query pairs\\
CareGPT~\cite{wang2023caregpt}                  &2023 &RLHF & Chinese and Western Medicine Corpus &HalluQA\\ 
ShenNong-TCM-LLM~\cite{zhu2023ChatMed}          &2023 &IFT  & Chinese Medicine Instruction Dataset&-\\ 
MedicalGPT~\cite{MedicalGPT}                    &2023 &PT, RLHF& Medical Dialogues&-\\ 
ChatMed~\cite{zhu2023ChatMed1}                   &2023 &IFT  & Online Consultation&-\\ 
QiZhenGPT~\cite{cmkrg_cmkrgqizhengpt_2024}      &2023 &IFT  &Pharmaceutical Knowledge\&A&-\\ 
Med-ChatGLM~\cite{ChatGLM-Med}                  &2023 &IFT  &Chinese Medical Knowledge Base&-\\ 
HuatuoGPT-II~\cite{chen2023huatuogpt}           &2023 &PT, IFT & Chinese and English Medical Texts &MMLU, CMMLU, C-EVAL, MedQA, MedMCQA\\ 
WiNGPT~\cite{research_winninghealthwingpt2_2024} &2023 &PT, IFT & Medical and General Knowledge&C-EVAL, WiNEval \\
Taiyi-LLM~\cite{luo_taiyi_2024}                  &2023 &PT, IFT & Biomedical Datasets &13 test sets including BC5CDR \\
Zhongjing \cite{yang2024zhongjing}&2023&PT, SFT,  RLHF&Medical Books, ChatMed, Medical Wiki& CMtMedQA, huatuo-26M\\
Med-Gemini ~\cite{saab_capabilities_2024}       &2024&IFT&Medical Knowledge, Clinical Cases & NEJM, USMLE-MM, MMMU-HM, ECG-QA\\
Health-LLM \cite{jin2024health}&2024&IFT&PMData, LifeSnaps, GLOBEM, AW\_FB& Health Prediction Tasks\\  \hline
\end{tabular}}
\end{table*}

\subsection{Datasets for Med-LLMs}
In this section, we collect and report some relevant datasets for Med-LLMs. As shown in Table \ref{llm_data}, we introduce the basic information of these datasets, including publication year, data scale, clinical task, and supporting languages. Besides, we also provided official links of these datasets to facilitate easier access for researchers. 

Developing high-quality datasets is one of the most crucial and important directions of Med-LLMs due to the challenges of accessing diverse medical datasets. Datasets of Med-LLMs play a fundamental role not only in providing large-scale training corpus with expert-level knowledge but also in conducting comprehensive and fair evaluations of LLM models towards specific tasks. The rapid development of LLMs in recent years has motivated a significant demand for large-scale, high-quality datasets, especially in the field of intelligent medical research.

\subsection{Evaluations for Med-LLMs}
\subsubsection{Quantitative Evaluations.} As shown in Table \ref{tab:nlp-eval-metrics}, different tasks are evaluated via various metrics, covering accuracy, F1 score, and perplexity. Here, we introduce the details of some important metrics.

$\bullet$ \textit{Perplexity.} Perplexity (PPL) is a standard that evaluates how well a probability model can predict a sample. When applied to language models like GPT, it represents the exponential average negative log-likelihood of a sequence. In essence, a lower perplexity score suggests that the model has a higher certainty in its predictions.
\begin{equation}
        \text{PPL} = 2^{-\frac{1}{n}\sum_{i=1}^{n} \log_2 P(w_i)},
\end{equation}
where $P(w_i)$ is the conditional probability of the model predicting the $i$-th word $w_i$ given the context of previous words. $n$ is the total number of words.

$\bullet$ \textit{ROUGE.} ROUGE scores compare the overlapping units (n-grams, skip-bigrams, or longest common subsequences) between the generated text and the reference. There are many ROUGE variants to measure the overlaps from different viewpoints, such as ROUGE-N, ROUGE-L, ROUGE-W, and ROUGE-SU. Here, we only give a detailed explanation for the most basic ROUGE-N. ROUGE-N is a metric that quantifies the overlapping N-grams between a candidate summary and a reference summary. Let $C$ be the set of all unigrams (or bigrams for ROUGE-2) in the candidate summary and $R$ be the set of all unigrams (or bigrams) in the reference summary.
ROUGE-1 can be calculated as:
\begin{equation}
\begin{aligned} 
 P_{\text{ROUGE-1}} &= \frac{\sum_{w \in C} \text{Count}{R}(w)}{|C|}, \\ 
 R_{\text{ROUGE-1}} &= \frac{\sum_{w \in C} \text{Count}{R}(w)}{|R|}, \\ 
 F_{\text{ROUGE-1}} &= \frac{(\beta^2 + 1) \cdot P_{\text{ROUGE-1}} \cdot R_{\text{ROUGE-1}}}{\beta^2 \cdot P_{\text{ROUGE-1}} + R_{\text{ROUGE-1}}}, 
 \end{aligned}
\end{equation}
where $\text{Count}{R}(w)$ denotes the frequency of the unigrams $w$ in the reference.

$\bullet$ \textit{BLEU.} BLEU calculates the similarity between a candidate translation and one or more reference translations based on n-gram overlap. The idea is that a good translation should contain many n-grams (sequences of words) that are also present in the reference translations. Mathematically, the BLEU score is computed as follows:
\begin{equation}
      \begin{aligned} 
    p_n &= \frac{\sum_{c\in C} \text{Count}_{clip}(ngrams(n,c))}{\sum_{c\in C} \text{Count}(ngrams(n,c))} \\
    \text{BP} &= \begin{cases} e^{(1 - \frac{r}{c})} & \text{if } c < r \\ 1 & \text{otherwise} \end{cases} \\ \text{BLEU} &= BP \cdot \exp\left(\sum_{n=1}^{N} w_n \log p_n\right) 
    \end{aligned}
\end{equation}
$p_n$ is the modified n-gram precision. $\text{BP}$ is the brevity penalty. $C$ is the set of candidate sentences. $\operatorname{Count}$ is to get all n-grams in the candidate sentence. $\text{Count}_{clip}$ counts clipped n-grams in the candidate that do not exceed the counts in any reference. $r$ and $c$ are the length of the reference and the candidate respectively. $N$ is the maximum n-gram order used for evaluation. $w_n$ are the weights assigned to each n-gram precision (usually uniform weights).

\subsubsection{Qualitative Evaluations}
Qualitative evaluations in NLP refer to non-numeric assessments that focus on the subjective and interpretive aspects of language understanding and generation tasks. Unlike quantitative metrics that rely on numerical measurements such as accuracy, precision, recall, or F1 scores, qualitative evaluations often involve human judgment to analyze the nuances of text produced by LLMs. Here are some common qualitative evaluation approaches:
\begin{itemize}[leftmargin=*]
    \item \textit{Human Evaluation.} Experts or human workers manually assess the generated text based on factors such as readability, coherence, grammar, style, relevance, and factual correctness. This can be done through rating scales or comparative studies between different systems.
    \item \textit{Error Analysis.} Researchers closely examine and categorize errors made by the system to understand its strengths and weaknesses, which helps inform improvements to the model or algorithm.
    \item \textit{Case Studies.} Detailed examination of specific examples to illustrate successes or failures of a system in specific contexts, which can reveal issues with edge cases or complexities that quantitative metrics might miss.
    \item \textit{User Feedback.} Collecting feedback from end-users interacting with LLM applications to understand the effectiveness and usability of the language outputs.
    \item \textit{Thematic Analysis.} To evaluate the LLMs' ability of effectively capturing context and conveying meaning, it is feasible to analyze themes and patterns that emerge in the text generated by the system 
    \item \textit{Aesthetic Judgments.} It is difficult to numerically quantify  the aesthetic qualities of generated text, such as creativity, emotional appeal, or narrative flow.
    \item \textit{Ethical and Societal Impact Assessment.} Examining the ethical implications and societal effects of an LLM system range from biases and fairness to privacy concerns.
    
\end{itemize}

\begin{table}[!t]
\centering
\caption{Commonly Used Evaluation Metrics for NLP tasks.}
\label{tab:nlp-eval-metrics}
\begin{tabular}{l|p{4.5cm}}
\hline
\textbf{Task} & \textbf{Evaluation Metrics} \\ \hline
Text Classification & \makecell[l]{Accuracy, Recall, Precision, F1 Score,\\ ROC Curve, AUC Value} \\ \hline
Text Generation & \makecell[l]{BLEU, METEOR, ROUGE, CIDEr,\\ Perplexity} \\ \hline
Machine Translation & \makecell[l]{BLEU, Translation Edit Rate, METEOR,\\ Adequacy, Fluency} \\ \hline
Question Answering Systems & \makecell[l]{EM (Exact Match), F1 Score, Mean\\Reciprocal Rank (MRR), Hits@k} \\ \hline
Sentiment Analysis & \makecell[l]{Accuracy, Macro-F1, Micro-F1,\\ Cohen's Kappa, Sentiment Scores} \\ \hline
Named Entity Recognition & \makecell[l]{Precision, Recall, F1 Score, Span-Level\\ Metrics, Entity Type Level Metrics}\\\hline
\end{tabular}
\end{table}
\subsubsection{Evaluation Benchmark for LLMs' Clinical Capabilities} The above evaluation methods heavily rely on intensive interactions with LLMs to obtain diagnostic dialogues, as well as expensive expert-level human labor to measure the diagnosis quality. Therefore, the automatic evaluation approaches gradually capture the research attention for assessing the practical clinical capabilities of LLMs.

$\bullet$ \textit{MedBench} \cite{cai2024medbench}. MedBench is a comprehensive evaluation benchmark for the medical domain, which consists of 40,041 questions about various branches of medicine including authentic examinations and clinical reports. The main components of MedBench involve: Chinese Medical Licensing Examination, Resident Standardization Training Examination, Doctor In-Charge Qualification Examination, and real-world clinic cases encompassing examinations, diagnoses, and treatments. Extensive experiments provided the following findings: (1) Chinese medical LLMs (\textit{e.g.}, HuaTuoGPT \cite{zhang2023huatuogpt}) underperform on the MedBench, indicating the importance of clinical knowledge and diagnostic precision. (2) Some general LLMs (\textit{e.g.}, GPT 4) surprisingly convey considerable clinical knowledge. 

$\bullet$ \textit{AutoEval} \cite{liao2023automatic}. AutoEval is an automated evaluation framework to evaluate the realistic capabilities of LLMs, \textit{i.e.}, working as virtual doctors for multi-turn dialogue. The clinical demands for consultation tasks require LLMs to (1) consider the factors they do not know, (2) query about missing medical factors of patients, and (3) generate diagnosis and treatment results. To this end, AutoEval is to reformulate USMLE into medical multiple-choice questions. Then, several metrics are designed to indicate the ability of doctor LLMs in multi-turn consultation, \textit{i.e.}, Medical Information Coverage Rate, Accuracy of the Final Task, Average Turn, and Average Length.

\begin{figure}[t]
\centering
\includegraphics[width=1\linewidth]{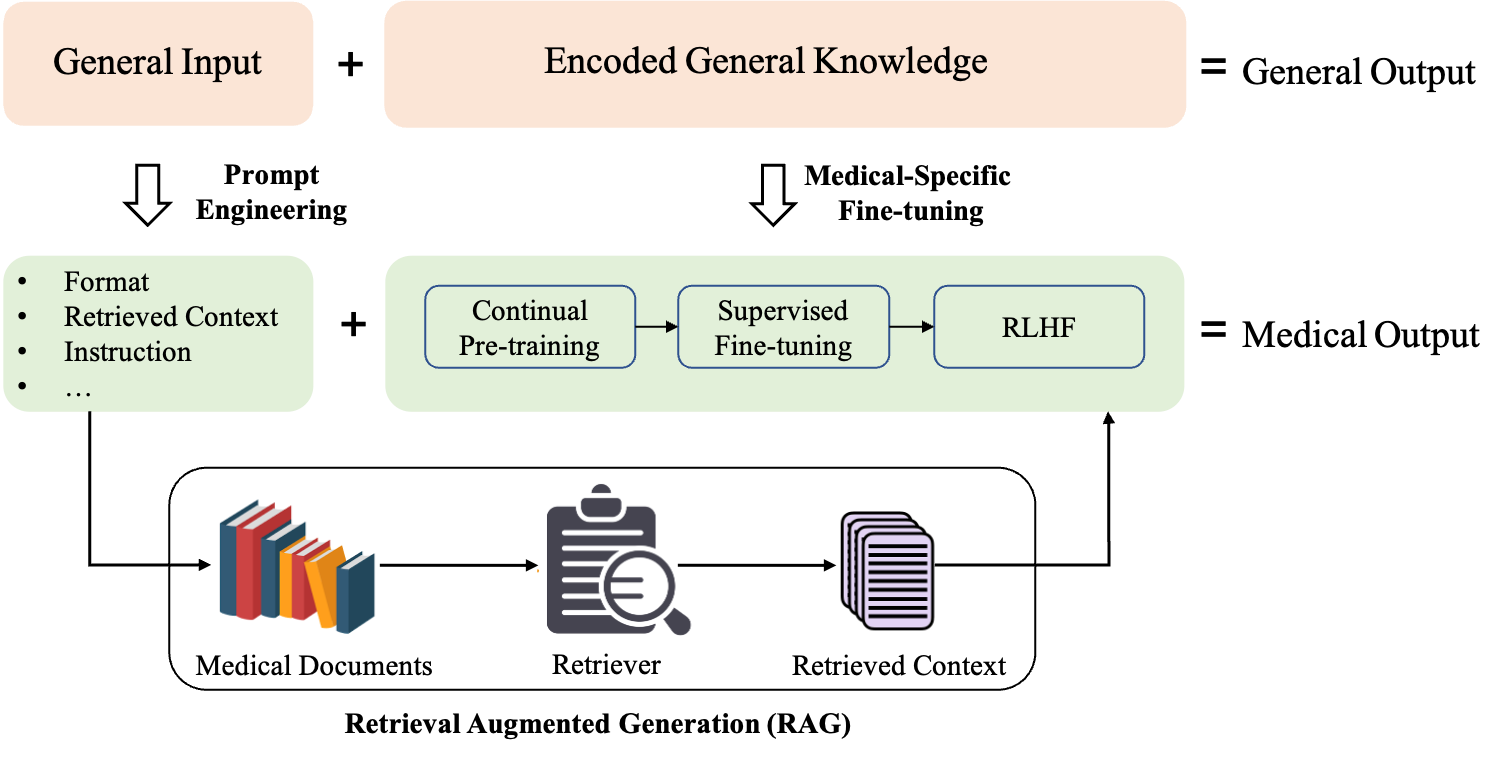}\\
\caption{Making a LLM to be a Doctor: A Multi-Step Approach. Prompt engineer can craft suitable prompts to derive the desired responses. Medical-specific fine-tuning can update parameters of a pre-trained LLM on a medical datasets to improve the clinical performance. RAG is to combining prompt engineering with context retrieval from external medical documents.
}
\label{figure:gen2med}
\end{figure}

$\bullet$ \textit{LLM-Mini-CEX} \cite{shi2023llm}. Based on MiniCEX \cite{norcini2003mini, kogan2002implementation}, an evaluation criterion called LLM-specific Mini-CEX is established to effectively assess the clinical capabilities of LLMs. Then, a patient simulator is developed to simulate the conversations with LLMs, where ChatGPT is utilized to automatically evaluate diagnosis dialogues.

$\bullet$ \textit{MedGPTEval} \cite{xu2023medgpteval}. MedGPTEval contains Chinese medical datasets and public benchmarks. The evaluation metrics are designed based on a comprehensive literature review, including medical professional capabilities, social comprehensive capabilities, contextual capabilities, and computational robustness, with 16 detailed indicators. These metrics have been optimized by a Delphi method using 5 experts in medicine and engineering \cite{xu2023medgpteval}. Then, 3 clinical experts manually construct a set of medical datasets for interactions, including 7 case reports and 27 medical dialogues.

$\bullet$ \textit{LLM-Human Evaluation} \cite{chiang2023can}. LLMs have exhibited superior performance on unseen tasks, even only trained on the specific-task instructions. The study \cite{chiang2023can} investigates that if such an ability can be utilized as an alternative to human evaluation. Both human and LLM evaluations are conducted for two NLP tasks: adversarial attacks and open-ended story generation. The experiments indicate that LLM evaluation exhibits consistent results with human-level evaluation. Besides, LLM evaluation provides more stable results over different task instructions and the sampling algorithms for generation.

$\bullet$ \textit{RJUA-SPs} \cite{liu2024towards}. The RJUA-SPs evaluation approach contains three basic elements: (1) Metric. Professional clinical practice pathways are exploited to define the clinical capabilities that a doctor should possess, named LLM-specific clinical pathway (LCP). (2) Data. Standardized Patients (SPs) are utilized to guide the data collection, which aims to well maintain the completeness of the evaluation. (3) Algorithm. A multi-agent framework is formulated to simulate the interactive environment between SPs and a doctor agent. A retrieval-augmented algorithm is specially designed to measure whether the behaviors of a LLM doctor are consistent with LCP.


\subsection{Making a LLM to be a doctor}
Transforming a general large language model (LLM) into a specialized medical LLM, akin to training it to become a ``doctor'', involves several sophisticated steps that go beyond simple parameter adjustments. As shown in Figure \ref{figure:gen2med}, the overall process contains a multi-step approach: prompt engineering, medical-specific fine-tuning, and RAG.

$\bullet$ \textit{Prompt Engineering}. This step involves crafting and refining prompts that can effectively guide the model to produce medically accurate and contextually appropriate outputs. The process begins with developing highly specific prompts, including detailed instructions, examples, and constraints that align closely with the medical domain. For instance, prompts might be designed to ask for diagnoses based on symptoms, treatment options for specific conditions, or explanations of complex medical procedures in layman's terms. It's crucial to incorporate elements such as patient history, current medications, and relevant clinical guidelines within the prompts to ensure the personalized responses.

$\bullet$ \textit{Medical-Specific Fine-tuning}. The fine-tuning begins with selecting a high-quality dataset that encompasses a wide array of medical knowledge, including clinical notes, textbooks, research articles, patient records (with proper anonymization), and other relevant materials. The data should cover various aspects of medicine such as pathology, pharmacology, anatomy, diagnostics, treatment protocols, and patient care practices. Ensuring diversity in the dataset helps the model generalize better across different medical scenarios and specialties. Moreover, the fine-tuning process should include continuous learning mechanisms that allow the LLM to stay updated with the latest medical research and guidelines. This can be achieved through periodic retraining with new data or by implementing online learning techniques where the model can learn from new inputs without forgetting previously learned information—a challenge known as catastrophic forgetting.

$\bullet$ \textit{Medical-Specific RAG}.
Additionally, implementing a medical RAG requires integrating robust retrieval mechanisms that allow the model to access and leverage up-to-date, evidence-based medical information. RAG enhances the LLM's capabilities by combining its generative power with precise, contextually relevant data retrieved from trusted medical databases and literature. This hybrid approach ensures that the outputs are not only coherent and contextually appropriate but also grounded in the latest clinical guidelines, research findings, and patient care standards. To achieve this, the system must be equipped with an efficient retrieval component capable of understanding complex medical queries and retrieving pertinent documents or passages from a vast corpus of medical knowledge. This corpus could include peer-reviewed articles, clinical trial results, textbooks, and practice guidelines. The retrieval process should prioritize high-quality sources and be able to handle nuanced queries related to diagnosis, treatment options, drug interactions, and patient-specific considerations.

Last but not least, rigorous testing through simulated clinical scenarios and peer review by medical professionals helps validate the model's efficacy and reliability in real-world applications. Through these methods, a general LLM can be effectively adapted to serve as a powerful tool for supporting healthcare providers and improving patient outcomes.

\subsection{Specific Med-LLMs}
Existing LLMs in the medical field, such as Med-PALM, Codex-Med, and MedAlpaca, have contributed to advancements in healthcare with their unique design objectives, architectures, and functionalities. As shown in Figure \ref{figure:timeline}, these models collectively drive the development of Med-LLMs and enhance their application in healthcare, providing robust support for future medical practice and research. Due to the limited space, we only simply introduce some Med-LLMs as follows.

\begin{itemize}[leftmargin=*]
    \item \textit{BioBERT}~\cite{lee2020biobert}: BioBERT is a variant of BERT specifically trained on biomedical text. It is designed to understand and process medical and biological information more effectively than general-purpose models. BioBERT is particularly useful for tasks such as named entity recognition, relation extraction, and text classification in the biomedical domain.
    \item \textit{PubMedBERT}~\cite{gu2021domain}: PubMedBERT is a BERT-based model trained on PubMed abstracts. It is particularly effective for tasks involving biomedical literature, such as literature review, information retrieval, and summarization. PubMedBERT helps researchers quickly sift through large volumes of scientific papers to find relevant information.
    \item \textit{ClinicalT5}~\cite{lu_clinicalt5_2022}: ClinicalT5 is a T5-based model trained on clinical data. It is designed to handle clinical text generation and summarization tasks. ClinicalT5 can generate detailed clinical reports, summarize patient records, and assist in clinical decision-making.
    \item \textit{GatorTron}~\cite{yang2022gatortron}: GatorTron is a large language model developed for various tasks, including those in the medical domain. It is known for its ability to handle complex medical text and generate high-quality outputs. GatorTron can be used for tasks such as generating detailed medical reports and summarizing clinical data.
    \item \textit{Codex-Med}~\cite{lievin2024can}: Codex-Med is a variant of the Codex model, specifically tailored for medical applications, which can assist in generating detailed medical reports and summarizing clinical data.
    \item \textit{Galactica}~\cite{taylor2022galactica}: Galactica is a large language model designed to handle a wide range of tasks, including those in the medical field. It is particularly effective for generating and understanding complex scientific and medical texts. Galactica can be used for literature review, information retrieval, and summarization in the biomedical domain.
    \item \textit{Med-PaLM}~\cite{singhal2023large}: Med-PaLM is a language model specifically designed for medical applications. It is trained on a large corpus of medical data and is used for tasks such as medical text generation and information extraction. Med-PaLM can generate detailed medical reports and assist in clinical decision-making.
    \item \textit{GPT-4-Med}~\cite{nori2023capabilities}: GPT-4-Med is a variant of the GPT-4 model, optimized for medical applications. It is designed to handle complex medical text and generate high-quality medical content. GPT-4-Med can be used for generating detailed medical reports, summarizing clinical data, and assisting in clinical decision-making.
    \item \textit{ChatDoctor}~\cite{yunxiang2023chatdoctor}: ChatDoctor is a conversational AI model designed to assist in medical consultations. It can answer medical questions and provide preliminary advice based on symptoms. ChatDoctor is useful for triaging patients and providing initial guidance before a formal medical consultation.
    \item \textit{BenTsao}~\cite{wang2023huatuo}: BenTsao is a language model designed for medical applications. It is trained to handle tasks such as medical text generation and information retrieval. BenTsao can generate detailed medical reports and assist in clinical decision-making.
    \item \textit{PMC-LLaMA}~\cite{wu2023pmc}: PMC-LLaMA is a language model trained on PubMed Central data. It is designed to handle tasks related to biomedical literature, such as text generation and information retrieval. PMC-LLaMA helps researchers quickly sift through large volumes of scientific papers to find relevant information.
    \item \textit{BianQue}~\cite{chenbianque}: BianQue is a language model designed for medical applications. It is trained to handle tasks such as medical text generation and information retrieval. BianQue can generate detailed medical reports and assist in clinical decision-making.
    \item \textit{Med-PaLM 2}~\cite{singhal2023towards}: Med-PaLM 2 is an updated version of Med-PaLM, designed to improve performance in medical applications. It is trained on a larger corpus of medical data and is used for tasks such as medical text generation and information extraction. Med-PaLM 2 can generate more accurate and detailed medical reports.
    \item \textit{GatorTronGPT}~\cite{peng2023study}: GatorTronGPT is a variant of GatorTron specifically designed for medical applications. It is used for tasks such as medical text generation and information retrieval. GatorTronGPT can generate detailed medical reports and assist in clinical decision-making.
    \item \textit{HuatuoGPT}~\cite{zhang2023huatuogpt}: HuatuoGPT is a language model designed for medical applications. It is trained to handle tasks such as medical text generation and information retrieval. HuatuoGPT can generate detailed medical reports and assist in clinical decision-making.
    \item \textit{Med-Flamingo}~\cite{moor2023med}: Med-Flamingo is a language model designed for medical applications. It is trained to handle tasks such as medical text generation and information retrieval. Med-Flamingo can generate detailed medical reports and assist in clinical decision-making.
    \item \textit{ShenNong-TCM-LLM}~\cite{zhu2023ChatMed}: ShenNong-TCM-LLM specializes in traditional Chinese medicine (TCM). This model is trained on extensive TCM texts, including ancient classics, modern research papers, and clinical practice guidelines. ShenNong-TCM-LLM supports practitioners in diagnosing and treating patients using TCM principles, helps in the formulation of herbal prescriptions, and aids in educating both practitioners and patients about TCM practices.
    \item \textit{MedicalGPT}~\cite{MedicalGPT}: MedicalGPT is a general-purpose medical language model designed to support a wide range of healthcare-related tasks. It is trained on diverse datasets that include clinical notes, medical textbooks, and public health resources. MedicalGPT can generate detailed medical reports, assist in clinical decision-making, and provide information for patient consultations. Its versatility makes it suitable for various roles within the healthcare system.
    \item \textit{QiZhenGPT}~\cite{cmkrg_cmkrgqizhengpt_2024}: QiZhenGPT focuses on traditional Chinese medicine diagnostics and treatment planning. It is trained on a comprehensive dataset of TCM case studies, diagnostic criteria, and therapeutic methods. QiZhenGPT assists TCM practitioners in making accurate diagnoses, recommending appropriate treatments, and educating patients about preventive health measures.
    \item \textit{Med-ChatGLM}~\cite{ChatGLM-Med} : Med-ChatGLM integrates conversational capabilities with deep medical knowledge. It is designed to engage in meaningful dialogues about health and wellness, provide personalized advice, and support users in managing chronic conditions. Med-ChatGLM can also assist healthcare providers in monitoring patient progress and adjusting treatment plans based on ongoing feedback.
    \item \textit{HuatuoGPT-II}~\cite{chen2023huatuogpt}: HuatuoGPT-II is an advanced version of HuatuoGPT, optimized for improved performance in medical applications. It leverages a larger training dataset and refined algorithms to enhance its ability to generate detailed medical reports, summarize clinical data, and assist in clinical decision-making. HuatuoGPT-II's improvements make it a valuable tool for healthcare professionals seeking precise and reliable AI support.
    \item \textit{Taiyi-LLM}~\cite{luo_taiyi_2024}: Taiyi-LLM is a language model specialized in traditional Chinese medicine. It is trained on historical texts, contemporary research, and clinical practice data. Taiyi-LLM supports TCM practitioners in diagnosing and treating patients, formulating herbal prescriptions, and educating patients about TCM theories and practices. Its expertise in TCM makes it a valuable resource for integrating traditional and modern medical approaches.
    \item \textit{Zhongjing}~\cite{yang2024zhongjing}: Zhongjing is a language model focused on traditional Chinese medicine, named after the renowned physician Zhang Zhongjing. It is trained on a rich corpus of TCM literature and clinical data. Zhongjing aids practitioners in making accurate diagnoses, recommending effective treatments, and educating patients about TCM principles. It serves as a bridge between traditional wisdom and modern healthcare practices.
    \item \textit{Med-Gemini}~\cite{saab_capabilities_2024}: Med-Gemini is a dual-language model capable of handling both English and Chinese medical texts. It is trained on a bilingual corpus of medical literature, clinical notes, and patient records. Med-Gemini facilitates cross-cultural medical communication, supports bilingual healthcare environments, and assists in translating medical documents accurately. Its dual-language capability enhances its utility in international healthcare settings.
\end{itemize}

\section{Improving Algorithms for Med-LLMs}
\label{sec:improve}
\subsection{Clinical Reasoning for Med-LLMs}
Clinical reasoning \cite{mattingly1991clinical} in the context of LLMs refers to the ability of these models to mimic and assist in the complex thought processes involved in medical diagnosis, treatment planning, and patient management, akin to how a human clinician would reason. Achieving advanced clinical reasoning capabilities in LLMs involves equipping them with an understanding of medical knowledge, and the ability to analyze patient data holistically, consider differential diagnoses, and make decisions based on the best available evidence. 

\subsubsection{General Algorithm} The term "Chain-of-Thought" refers to a cognitive process where an individual mentally links a series of thoughts, ideas, or reasoning steps to solve a problem, make a decision, or understand a concept. This can include breaking down a question into sub-problems, considering possible solutions, eliminating incorrect options based on available knowledge, and justifying the final answer. By generating intermediate steps and explanations, LLMs become more transparent in their decision-making, facilitating understanding and trust.

For instance, a chain-of-thought process for Med-LLMs would involve breaking down complex medical scenarios, diagnoses, or treatment plans into a logical sequence of thoughts, leveraging the model's understanding of medical knowledge and context. To diagnose diseases based on symptoms, the reasoning path may be as follows: acknowledging the reported symptoms, cross-referencing them with known disease symptomatology from medical databases, weighing the likelihood of each potential diagnosis given patient demographics and medical history, and finally presenting the most probable diagnosis with supporting evidence and a confidence level. 

\subsubsection{Specific Reasoning Techniques} There are some other specific pathways to improve the clinical reasoning capabilities of Med-LLMs.

$\bullet$ \textit{In-Context Padding (ICP)} \cite{wu2024guiding}. ICP consists of four major steps to improve the reasoning capacity of LLM in the context of clinical environments: (1) extracting medical entities from the clinical context and the reasoning objective; (2) inferring relevant medical entities from KG; (3) concatenating the acquired knowledge seeds with the prompt; (4) generating the reasoning results as well the clinical explanation.

$\bullet$ \textit{JMLR} \cite{wang2024jmlr}. Unlike previous RAG methods where the retrieval model is trained separately from the LLM, JMLR \cite{wang2024jmlr} jointly trains the LLM and information retrieval model during the fine-tuning stage. A synchronized training mechanism is formulated to retrieve clinical guidelines and leverage medical knowledge, significantly improving the clinical reasoning capabilities of LLMs and decrease the requirements for computational resources.

\subsection{Knowledge Graph for Med-LLMs}
Despite the excellent capacities, LLMs often suffer from challenges on knowledge-intensive tasks, such as the potential to generate hallucinated content and the lack of domain-specific knowledge \cite{chu2024sora}. As a promising solution, knowledge graphs (KGs), which store enormous knowledge in the triple format can be utilized to enhance the task performance of LLMs by providing precise and necessary knowledge \cite{wang2024llmrg,wang2023enhancing,chu2024llm,sheu2021knowledge}.

\subsubsection{General Algorithms}
Generally, knowledge-enhanced approaches can be expanded into other data structures (\textit{e.g.}, tables and databases), while we limit our discussion to KG-enhanced LLMs. The roadmap of unifying KGs and LLMs includes KG-enhanced LLMs, LLM-augmented KGs, and Synergized LLMs + KGs \cite{pan2024unifying}. The KG-enhanced LLMs and LLM-augmented KGs are two dual frameworks that aim to enhance the capabilities of LLMs and KGs, respectively. Building upon these frameworks, Synergized LLMs + KGs is a unified framework to synergize LLMs and KGs to mutually enhance each other. In this work, we discuss a different aspect of the integration of KG and LLMs, namely Inference-Time LLM and Training-Time Augmentations.

$\bullet$ \textit{Inference-Time KG Augmentation.} In this approach, an LLM is augmented with a retrieval mechanism that fetches relevant information from a KG based on the context of the input \cite{baek2023knowledge,tian2024graph,wei2024llmrec}. The model can then use this external knowledge to inform its output \cite{wu2023retrieve}, reducing the likelihood of hallucinations and enhancing the domain-specific accuracy of its responses \cite{guan2024mitigating}. This is akin to giving the model a dynamic cheat sheet of verified facts to reference during its processing.

$\bullet$ \textit{Training-Time KG Augmentation.} This strategy goes beyond simple retrieval by creating a more integrated system where the LLM and KG interact more intimately \cite{li2023beginner,yang2023chatgpt}. It might involve training the LLM on tasks that explicitly leverage KG data, allowing the model to learn how to utilize structured knowledge more effectively within its natural language processing capabilities. This could include updating the model's parameters based on KG interactions, essentially teaching it how to reason with structured data.

\subsubsection{Specific KG-Augmented Med-LLMs} The LLMs' performance is significantly improved in tasks related to healthcare, medicine, and biomedical research by providing it with a rich, structured repository of medical knowledge. Here are some recent applications for KG-augmented LLM systems.

$\bullet$ \textit{DR.KNOWS} \cite{gao2023leveraging}. Guided by clinical diagnostic standards, DR.KNOWS is introduced as an innovative approach to improve LLMs in generating diagnostic outcomes. This method combines the utilization of a medical knowledge graph and the Unified Medical Language System (UMLS) of the National Library of Medicine. DR.KNOWS serves as an interpretive and summarizing aide, harnessing the medical KG to untangle complex medical notions. Tt establishes a clear, justifiable route to diagnoses, facilitating the employment of AI-augmented decision support mechanisms that offer transparency in their analytical processes. Consequently, this integration advances the precision of AI diagnoses while ensuring they remain comprehensible and accountable.

$\bullet$ \textit{KG-Rank} \cite{yang2024kg}. To tackle the issues of factual inconsistencies and inherent biases in medical inquiries, the KG-Rank framework \cite{yang2024kg} is introduced as an enhancement to LLMs. This framework employs a medical knowledge graph augmented with ranking and re-ranking strategies, specifically designed to enhance the accuracy of free-text medical question-answering. Concretely, for a specific question, the triplets are initially retrieved from the medical KG to gather factual information. Subsequently, ranking algorithms are applied to refine this initial retrieval results, ordering the results in a manner that facilitates the generation of more precise responses. According to the report \cite{yang2024kg}, KG-Rank pioneers the integration of ranking models with KGs in the context of medical question answering for producing elaborate answers.

$\bullet$ \textit{MedKgConv} \cite{varshney2023knowledge}. Existing generative dialog approaches may suffer from monotonous and uninteresting conversations. To address this issue, MedKgConv \cite{varshney2023knowledge} proposed to combine various pre-trained language models with UMLS to generate human-like conversations based on MedDialog-EN dataset. UMLS contains diverse medical-related information, \textit{i.e.}. disease, symptoms, and laboratory tests. To apply semantic information from the graphs, the reasoning step is conducted over the retrieved KG by reading the triples in each graph using MedFact attention. Then, a policy network is used to effectively inject relevant entities into the response text.

$\bullet$ \textit{ChiMed} \cite{ye2023qilin}. The Chinese medicine dataset is built to train the Qilin-med model, including medical question answering, plain texts, knowledge graphs, and dialogues. The knowledge graph subset contains the data from CPubMed-KG \cite{xiang_cpubmed_kg}, 39Health-KG \cite{chen2018qasy}, and Xywy-KG \cite{bai2019chatbot}. Diverse medical information is included to ensure the KG comprehensiveness, \textit{e.g.}, causation, symptoms, and recommended drugs.

$\bullet$ \textit{DISC-MedLLM} \cite{bao2023disc}. To construct high-quality SFT datasets, DISC-MedLLM \cite{bao2023disc} proposed a medical KG-driven sample construction to generate accurate and truthful medical responses. A department-oriented strategy is utilized to select triples from a medical KG according to the patient queries. For each triple, GPT-3.5 is exploited to generate QA pairs in a few-shot manner.

\subsection{LLM-based Medical Agents}
Inspired by the cutting-edge evolution of AI agents, LLMs are not just utilized for generating text but act as central systems for autonomous agent systems \cite{talebirad2023multi,chu2024professional,guan2023intelligent}. This paradigm represents a significant leap forward in AI capabilities, moving from passive question-answering systems to proactive, versatile agents capable of executing complex tasks.

\subsubsection{General Algorithm} In general, a typical agent system should consist of the following components:

\begin{itemize}[leftmargin=*]
    \item Planning Component \cite{zhu2024knowagent}. The planning component within these AI agents is pivotal for strategic thinking. It empowers the agent to deconstruct multifaceted objectives into simpler, actionable steps. By doing so, the agent can systematically navigate through a task, adjusting its course as needed based on intermediate results. This hierarchical task breakdown is akin to how humans tackle complex problems, enhancing the agent's efficiency and adaptability.
    \item Memory Component \cite{hou2024my}. Memory is another cornerstone of these autonomous systems. It goes beyond short-term context retention seen in earlier chatbots and incorporates long-term storage and retrieval mechanisms. The integration of external vector databases ensures that agents can accumulate, organize, and access a vast corpus of information, simulating a persistent knowledge base. This memory aids in decision-making, allowing the agent to learn from past experiences and apply that learning to new situations.
    \item Tool Utilization \cite{mei2024llm}. With planning and memory in harmony, AI agents can interact dynamically with their environment. They can leverage various tools and execute actions, mirroring human agency, such as autonomously devising strategies and implementation plans for given topics, and showcasing advanced planning, memory utilization, and action execution capabilities.
    \item Evaluation \cite{mehandru2024evaluating}. The advent of benchmarks like AgentBench \cite{liu2023agentbench} is crucial for systematically evaluating the performance of LLMs in agent-like roles. By presenting a spectrum of tasks from web navigation to game playing and knowledge management, it pushes the boundaries of AI research, encouraging the development of more sophisticated and adaptable agents.
\end{itemize}

These components pave the way for AI agents that are not only more autonomous but also better equipped to understand and interact with the world around them. For instance, the LangChain library \cite{yao2022react} is a framework for developing LLM-based applications, which simplifies the entire practical application lifecycle for integrating LLMs with broader computational ecosystems. The building agent systems can tap into diverse data sources and computational tools, facilitating the creation of more AI solutions.

\subsubsection{Specific Medical Agent} Here, we introduce some LLM-based medical agents. These agents should complement rather than replace human healthcare providers, ensuring that the nuanced understanding and empathy that humans bring to patient care are not lost.

$\bullet$ \textit{CT-Agent} \cite{yue2024ct}. Motivated by the advanced LLMs and multi-agent systems, CT-Agent (Clinical Agent System) is introduced as an integrated approach to improve accessibility and utility for clinical tasks. CT-Agent leverages GPT-4 \cite{achiam2023gpt}, multi-agent architectures, LEAST-TO-MOST \cite{zhou2022least}, and ReAct \cite{yao2022react} reasoning technology, which can boost the performance in clinical contexts for managing the entire clinical process.

$\bullet$ \textit{AutoGen} \cite{wu2023autogen}. AutoGen is an open-source and customizable framework for building effective and efficient applications, including mathematics, coding, QA, and online decision-making. Based on AutoGen, users can design multi-agent applications in a convenient manner, where interactions among them are conducted to finish final tasks. AutoGen can work in various modes via the combinations of LLMs, human inputs, and tools. As stated in \cite{wu2023autogen}, both natural language and computer code can be applied to formulate conversation patterns. Therefore, AutoGen-based users can flexibly define agent behaviors. 

$\bullet$ \textit{ArgMed-Agents} \cite{hong2024argmed}. To satisfy the requirements of clinical tasks (\textit{e.g.}, complex reasoning and planning), a multi-agent framework called ArgMed-Agents is proposed to provide explainable clinical decision reasoning. ArgMed-Agents is able to mimic the procedure of clinical reasoning by generating explanations in a self-directed way. The core is to perform self-argumentation iterations based on a reasoning mechanism for modeling cognitive processes. Then an augmented process is taken as a directed graph to represent conflicting relationships. Then, a symbolic solver can identify rational and coherent arguments for the decision.
 
$\bullet$ \textit{MAD} \cite{smit2023we}. Multi-agent debate (MAD) \cite{smit2023we} is proposed as a useful strategy to enhance the truthfulness of LLMs, which explores various debating and prompting strategies to maintain the trade-offs among cost, time, and accuracy. MAD protocols are difficult to optimize due to their sensitivity to hyperparameters. Some insights are proposed to improve debating strategies, such as adjusting agent agreement levels.

\subsection{Retrieval-Augmented Generation for Med-LLMs}
Retrieval-Augmented Generation (RAG) \cite{gao2023retrieval} is a machine learning technique that combines the strengths of retrieval-based and generative models to enhance the quality and diversity of generated text. This approach has gained significant attention in NLP tasks, particularly in areas like conversational AI, question-answering systems, and text summarization.

\subsubsection{General Algorithm} In general, a typical RAG system should consist of the following components:

\begin{itemize}[leftmargin=*]

\item Retrieval Component: The system starts by retrieving relevant information from a large database or corpus. This could involve indexing and efficiently searching through past conversations, documents, or web pages based on the input query. Techniques like TF-IDF \cite{ramos2003using}, BM25 \cite{trotman2014improvements}, or more advanced retrieval methods can be used for this component.

\item Generation Component: Once the relevant information is retrieved, a generative model (\textit{e.g.}, GPT-3, T5, or BERT) can use this information as context to generate a response or output \cite{izacard2023atlas}. The generation process is augmented by conditioning the model on the retrieved data, allowing it to generate more informed, contextually accurate, and diverse responses rather than generating from scratch.

\item Component Integration: The integration ways of these two components can vary \cite{gao2023retrieval}. Firstly, the retrieved information might be concatenated with the input prompt and directly fed into the generator. Others might use a more sophisticated fusion mechanism, where the retrieval and generation models interact in multiple steps, refining the context and the generated output iteratively.
\end{itemize}

By leveraging external knowledge, the generated text of RAG is more likely to be contextually appropriate and accurate. Retrieval of varied sources can introduce more diversity in the generated outputs, reducing the likelihood of repetitive or generic responses \cite{lewis2020retrieval}. Retrieval models can quickly narrow down the scope of information needed, which can make the generation process more efficient compared to exploring the entire knowledge space. Incorporating specific retrieved information can provide more control over the content and tone of the generated text, aligning it better with user expectations or specific requirements. 

\subsubsection{Specific RAG Algorithm} We summarize some RAG applications in the medical domain. By integrating external knowledge with generative models, RAG can provide more accurate, up-to-date, and contextually relevant information compared to traditional methods.

$\bullet$ \textit{Clinfo.ai} \cite{lozano2023clinfo}. Clinfo.ai is an open-source WebApp that answers clinical questions based on dynamically retrieved scientific literature. The information retrieval and summarization tasks are applied to evaluate the retrieval-augmented LLM systems.

$\bullet$ \textit{Almanac} \cite{zakka2024almanac}. Incorrect and harmful generations indeed limit the clinical applications of LLMs. To address this issue, Almanac is equipped with retrieval capabilities from curated medical resources for clinical guidelines and treatment recommendations. A group of clinicians and health care practitioners are utilized to evaluate Almanac, which makes a comparison between the responses from Almanac and standard LLMs (\textit{e.g.}, ChatGPT-4).

$\bullet$ \textit{BiomedRAG} \cite{li2024biomedrag}. Different from previous retrieval-augmented LLMs, BiomedRAG  \cite{li2024biomedrag} exploits a straightforward RAG method by directly inputting the retrieved documents into the LLM, rather than utilizing cross-attention mechanisms to encode retrieved text. BiomedRAG can effectively decrease the negative effects of noise information in retrieved documents, particularly for noise-intensive tasks. The study demonstrated the effectiveness of LLMs working as the supervised signal for the retrieval model in the biomedical domain.

$\bullet$ \textit{Self-BioRAG} \cite{jeong2024improving}. To improve RAG generalization for domain-specific problems, the Self-BioRAG framework \cite{jeong2024improving} is introduced for biomedical text, which specializes in explanation generation, domain-specific document retrieval, and self-reflecting generated responses. 84k biomedical instruction sets are exploited to train Self-BioRAG. The study indicated that domain-specific components, such as a retriever, domain-related document corpus, and instruction sets are necessary for adhering to domain-related instructions.

$\bullet$ \textit{ECG-RAG} \cite{yu2023zero}. To investigate LLMs for the ECG diagnosis, a zero-shot retrieval-augmented diagnosis technique \cite{yu2023zero} is designed to utilize the inherent encoded knowledge while infusing expert knowledge for carefully crafting prompts. Datasets of specific domain knowledge are built filled with cardiac symptoms and sleep apnea diagnosis.

$\bullet$ \textit{ChatENT} \cite{long2023chatent}. Existing LLMs suffer from inherent limitations including inconsistent accuracy, specific prompting requirements, and the risk of generating harmful hallucinations \cite{he2023survey}. A domain-specific fine-tuned model would effectively address these limitations. OHNS-relevant data is gathered from open-access internet sources. Retrieval-Augmented Language Modeling (RALM) \cite{long2023chatent} is utilized to recall this information for pre-training, which is integrated into ChatGPT to create an OHNS-specific knowledge QA platform, named as ChatENT.

$\bullet$ \textit{MIRAGE} \cite{xiong2024benchmarking}. Medical Information Retrieval-Augmented Generation Evaluation (MIRAGE) \cite{xiong2024benchmarking} is constructed to evaluate the medical RAG systems, including 7,663 questions from five medical QA datasets. MEDRAG can improve the performance of various LLMs, outperforming Chain-of-Thought prompting, GPT-3.5, Mixtral, and GPT-4 .

$\bullet$ \textit{MedicineQA} \cite{huang2024tool}. Due to the lack of domain-specific knowledge, it is challenging to deploy LLM models into medical scenarios. A multi-round dialogue benchmark called MedicineQA  \cite{huang2024tool} is proposed to simulate the real-world medication scenarios, requiring LLMs to answer with retrieved evidence from the medicine datasets. MedicineQA contains 300 multi-round question-answering pairs. Then a \textit{Distill-Retrieve-Read} framework is designed to replace previous \textit{Retrieve-then-Read}, which utilizes a tool calling mechanism to formulate search queries.

\subsection{Human Alignment for Med-LLMs}
\subsubsection{General Algorithm} LLMs trained on extensive textual corpora have emerged as leading solutions for a broad array of NLP tasks. Despite their notable performance, these models are prone to certain limitations such as misunderstanding human instructions, generating potentially biased content, or factually incorrect (hallucinated) information. Hence, aligning LLMs with human expectations has become an active area of interest within the research community \cite{tian2023chimed}. 
\begin{itemize}[leftmargin=*]
\item \textit{Data}. The core is to effectively collect large-scale and high-quality data for LLM alignment, leveraging NLP benchmarks, human annotators, and powerful LLMs to generate training instructions. 
\item \textit{Training}. Training methodologies involve optimizations for better efficiency and stability in incorporating human preferences, such as parameter-efficient training methods. Additionally, some studies consider human preference as ranking-based training signals or replace scalar rewards with language-based feedback to enhance training stability and performance.
\item \textit{Evaluation}. Various human-centric LLM evaluation benchmarks and automatic evaluation protocols have been proposed to obtain a comprehensive evaluation of aligned LLMs.
\end{itemize}

\subsubsection{Specific Alignment Algorithm} Achieving human alignment in medical LLMs is crucial because it directly impacts patient safety, trust in technology, and the overall effectiveness of healthcare delivery. Here are some human alignment algorithms, which can be approached in the development and deployment of medical LLMs.

$\bullet$ \textit{Safety Alignment} \cite{han2024towards}. The research carries out the first safety evaluation for Med-LLMs. To this end, the study \cite{han2024towards} summarizes the definitions of medical safety and alignment for medical AI systems. Then an evaluation dataset is constructed with harmful medical questions as indicators, which evaluates both general and medical safety and alignment of Med-LLMs.

$\bullet$ \textit{SELF-ALIGN} \cite{sun2024principle}. Due to the expensive costs of collecting human supervision and the quality issues (\textit{e.g.}, reliability, diversity, self-consistency, and undesirable biases), a SELF-ALIGN approach is proposed to combine principle-driven reasoning and the generative power of LLMs \cite{sun2024principle}, which aims to achieve the self-alignment with minimal human supervision. SELF-ALIGN contains four steps: (1) synthetic prompt generation to improve prompt diversity; (2) human-written principles for response generation; (3) high-quality self-aligned fine-tuning; (4) a refinement step to avoid simple or indirect responses.

$\bullet$ \textit{EGR} \cite{manathunga2023aligning}. It is effective for few-shot prompting to combine diverse techniques, which can significantly enhance the performance of LLMs. Motivated by this, the EGR (expand-guess-refine) alignment strategy \cite{manathunga2023aligning} is proposed for medical QA as a parameter and data-efficient solution.

\subsection{Multi-Modal Learning} 
A Multi-Modal Large Language Model (MM-LLM) is an advanced artificial intelligence system that integrates and processes multiple types of data, typically including text, images, audio, and sometimes video or other sensor data. These models extend beyond conventional language models, which primarily deal with text, by incorporating and understanding context from different sensory inputs.

\subsubsection{General Algorithms}  The core objective of MM-LLMs is to establish a unified framework for processing and generating content across various modes of communication, enabling more comprehensive and contextually rich interactions. There are several examples of general-purpose MM-LLMs:

$\bullet$ \textit{PaLM-E} \cite{driess2023palm}. PaLM-E is part of Google's Pathways initiative, which aims to create AI systems that can perform a wide range of tasks across different modalities. It builds upon the success of PaLM, a large-scale text-only language model, by integrating multimodal capabilities. PaLM-E incorporates both textual and visual embeddings, allowing it to understand and generate content in response to images and text prompts. It can perform image captioning, visual question answering, and even complex reasoning about visual scenes, all without requiring extensive fine-tuning on task-specific datasets.

$\bullet$ \textit{LLaVA} \cite{liu2024visual}.  The Large Language and Vision Assistant (LLaVA) model is introduced to bridge vision and language understanding through a multi-modal approach. It combines a vision encoder (CLIP \cite{radford2021learning}) with a language decoder (Vicuna \cite{chiang2023vicuna}) to form a general-purpose MM-LLM for vision-to-language tasks. The training process involves fine-tuning the decoder on 158,000 language-vision instruction examples from the MS-COCO dataset \cite{lin2014microsoft}.

$\bullet$ \textit{mPLUG-OWL} \cite{ye2023mplug}. mPLUG-OWL is first introduced by the researchers from the Alibaba DAMO academy, which could lead to inadequate alignment due to limited parameter flexibility. mPLUG-OWL utilizes advanced pre-trained LLM LLaMA-7B \cite{touvron2023llama} as the language decoder, while ViT-L/14 serves as the visual foundation model for the visual encoder to extract visual features from the input images. The ViT is initialized from the pre-trained CLIP ViT-L/14 model \cite{radford2021learning} for faster convergence.

\subsubsection{Specific Multi-Modal (Medical) LLMs} The literature \cite{hartsock2024vision} provides a more detailed discussion of the applications of vision-language models in the medical domain, including the exploration of medical vision-language datasets, in-depth analyses of architectures and pre-training strategies employed in recent noteworthy medical VLMs. Here, we only discuss some recent works for medical multi-modal LLMs.

$\bullet$ \textit{AD-MM-LLM} \cite{feng2023large}. For diagnosing Alzheimer's disease (AD), the study \cite{feng2023large} introduced a pre-trained LLM to non-image data for knowledge embedding and a ConvNeXt for image data. A multi-modal alignment is conducted for multi-level multi-modal feature fusion.

$\bullet$ \textit{RAMM} \cite{yuan2023ramm}. RAMM is a retrieval-augmented pretrain-and-finetune paradigm \cite{yuan2023ramm}, which could alleviate the data limitation issue for biomedical VQA. A new biomedical dataset named PMCPM is constructed by extracting image-text pairs from PubMed. Image-text contrastive objective (ITC) is exploited for pre-training. A retrieval-augmented method is to retrieve similar image-text pairs based on ITC from pre-training datasets.

$\bullet$ \textit{LLaVA-Med} \cite{li_llava-med_2023}. LLaVA-Med exhibits excellent multi-modal conversational capability. LLaVA-Med \cite{li_llava-med_2023} is required to align biomedical vocabulary based on figure-caption pairs and learn open-ended semantics using generated instruction-following data, which can follow open-ended instructions to assist with queries about a biomedical image. The training data includes a biomedical figure-caption dataset extracted from PubMed Central. Besides, LLaVA-Med can be trained in less than 15 hours using 8 A100s.

$\bullet$ \textit{Qilin-Med-VL} \cite{liu2023qilin}.  Qilin-Med-VL is the first Chinese large vision-language model designed for text and visual data, which relies on a pre-trained vision transformer with a foundational LLM. The training dataset is ChiMed-VL with more than 1M image-text pairs. A two-stage curriculum training process is conducted including feature alignment and instruction tuning, which improves the ability to generate medical captions and answer complex medical queries.


\begin{figure}[t]
	\centering
	\includegraphics[width=0.9\linewidth]{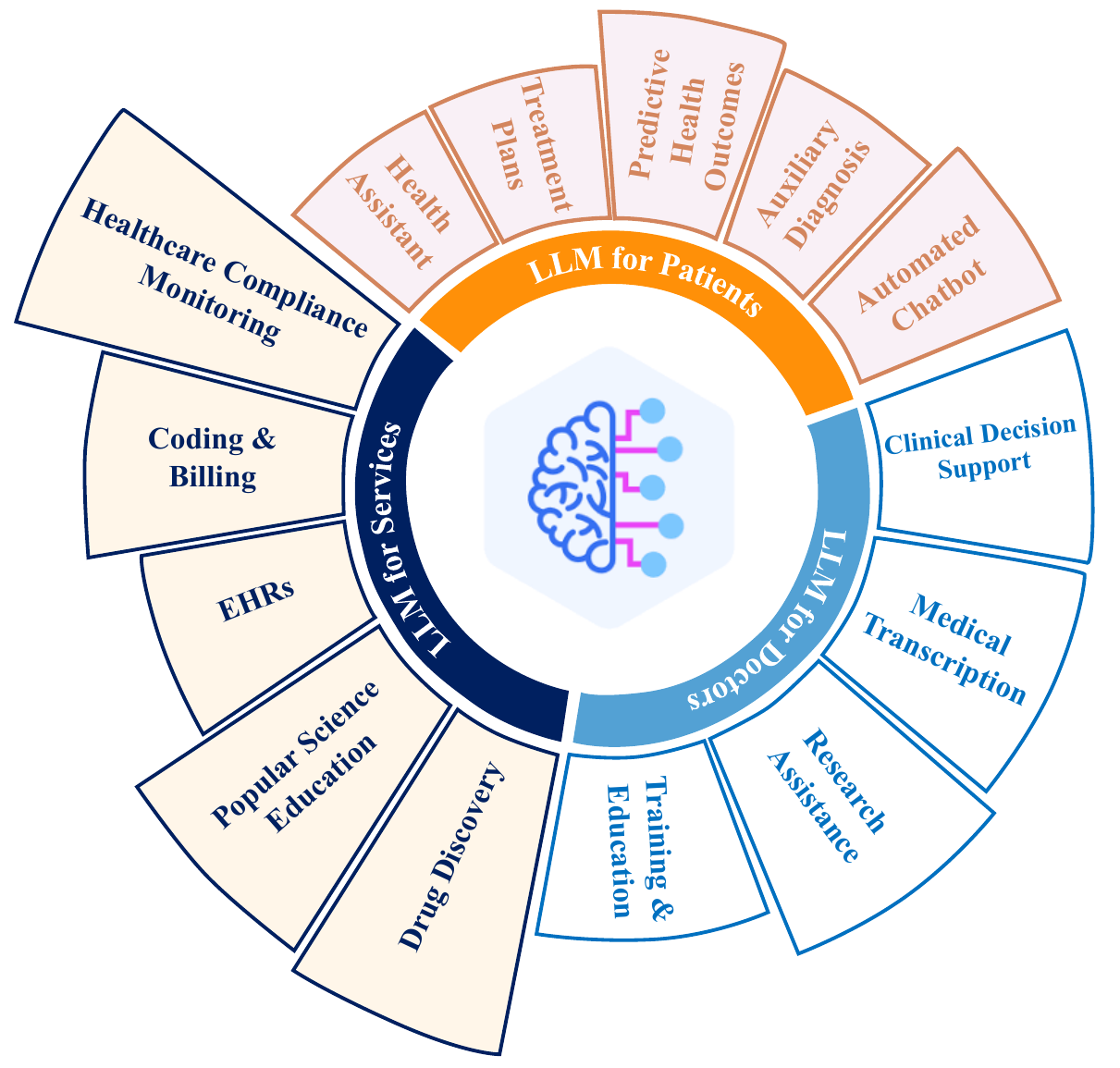}\\
	\caption{Applications of Med-LLMs. They assist in diagnosing illnesses, developing personalized treatment plans, analyzing medical records for pattern detection, supporting medical education training that offers professional advice and education, and powering intelligent robotics.}
	\label{figure:apply}
\end{figure}

\section{Applying Medical LLMs}
\label{sec:apply}
This section delves into a comprehensive investigation of the multifarious applications of LLM in the medical domain. Med-LLMs are experiencing rapid growth in the clinical context, medical education, and medical research, including clinical decision, medical record analysis, patient engagement, health information dissemination, \textit{etc}. 

\subsection{Multifarious Applications}

$\bullet$ \textit{Clinical Decision Support.} Med-LLMs can play a crucial role in enhancing clinical decision support systems \cite{wu2023medical}, which are designed to help healthcare professionals make informed decisions about patient care. Here are some specific ways Med-LLMs contribute to clinical decision support: (1) Symptom Analysis \cite{wang2024beyond}: By understanding natural language descriptions of symptoms entered by healthcare providers or patients, Med-LLMs can analyze and correlate these symptoms with a vast database of medical knowledge to suggest possible diagnoses. This can be particularly helpful in the identification of complex or rare conditions. (2) Risk Assessment: Med-LLMs can assess patient data, including demographics, medical history, lifestyle factors, and current health status, to assess the risk of developing certain conditions or complications. This enables proactive interventions and personalized prevention strategies \cite{chu2024task}. (3) Treatment Recommendations \cite{li2023meddm,chu2023leveraging,chu2020matching}: Based on the diagnosed condition, Med-LLMs can propose evidence-based treatment options, considering factors such as patient allergies, medication interactions, and the latest clinical guidelines. This supports clinicians in selecting the most appropriate therapy for each patient.

Various studies have clarified the broad prospects for the applications of LLMs in clinical decision support, \textit{e.g.}, radiology \cite{wang2023chatcad}, and oncology \cite{sorin2023large}. This synergy of AI technology with clinical expertise promises a paradigm shift in the quality and efficiency of medical decision-making processes.

$\bullet$ \textit{Clinical Report Generation.} Med-LLMs are also being leveraged for the generation of clinical reports \cite{chen2024dia}, streamlining documentation workflows and improving the accuracy and consistency of medical reporting. Med-LLMs can provide a flexible way for clinical report generation: (1) Automated Report Writing \cite{goel2023llms}: By processing input data from various sources such as electronic health records (EHRs), imaging study results, and laboratory tests, Med-LLMs can generate comprehensive, structured clinical reports automatically. This includes discharge summaries, radiology reports, pathology reports, and operative notes, saving time for healthcare professionals. (2) Customizable Templates \cite{ma2023impressiongpt}: Med-LLMs can incorporate institutional or specialty-specific templates, ensuring that generated reports follow the desired format and include all necessary information. This customization enhances the utility and readability of the reports for different clinical contexts. (3) Consistency and Accuracy: By relying on a standardized knowledge base,  Med-LLMs can reduce variability in language and improve the overall quality of reports. They ensure adherence to clinical standards and minimize the risk of errors or inconsistencies that can arise from manual reporting.

$\bullet$ \textit{Medical Education Training.} Med-LLMs are transforming medical education by offering innovative ways to teach, learn, and interact with medical knowledge \cite{safranek2023role}. Several key directions of Med-LLMs in the field of medical education involve (1) Question-Answering Systems: Students and practitioners can pose complex medical questions to these models, receiving accurate and detailed explanations drawn from a wide range of medical literature. This fosters a deeper understanding of concepts and encourages critical thinking. (2) Simulation of Clinical Cases \cite{liu2024towards}: Med-LLMs can simulate virtual patients with specific symptoms, medical histories, and test results, enabling learners to practice diagnostic reasoning and treatment planning in a safe, controlled environment \cite{chu2021graph,chu2023continual}. (3) Summarization of Research Papers \cite{van2024adapted}: Given the vast amount of medical literature published daily, LLMs can summarize key findings, methodologies, and implications from research papers, helping students stay updated with the latest advancements efficiently.

$\bullet$ \textit{Medical Assistive Robotics.} The integration of Med-LLMs with Medical Robotics is an emerging field that promises to revolutionize surgical training, patient care, and the development of advanced robotic systems. Here are some key areas where Med-LLMs can contribute to medical robotics: (1) Patient-Specific Surgical Planning \cite{xu2024enhancing}: With access to a patient's medical records and imaging data, Med-LLMs can assist in creating personalized surgical plans, taking into account the unique anatomy and health status of each patient. This information can guide robotic systems in executing highly customized surgeries. (2) Robotic Surgical Training \cite{chen2024llm}: By generating realistic surgical scenarios and providing real-time feedback, Med-LLMs can enhance the training of surgeons using robotic platforms. They can simulate various complications and patient responses, creating a more comprehensive and dynamic learning experience.

$\bullet$ \textit{Medical Language Translation.}
Med-LLMs can overcome linguistic barriers to enhance communication and improve healthcare delivery across diverse populations \cite{garcia2024medical}. Here's how Med-LLMs are revolutionizing the medical language translation task: (1) Precision in Medical Terminology: Trained on extensive medical corpora, these models grasp the intricacies of medical vocabulary, ensuring translations are not just linguistically accurate but also contextually precise. This is crucial for avoiding misunderstandings that could lead to misdiagnosis or incorrect treatment. (2) Real-Time Translation \cite{koshkin2024transllama}: In clinical settings, Med-LLMs can facilitate real-time conversations between healthcare providers and patients speaking different languages. This live translation capability streamlines communication during consultations, emergency situations, or remote care scenarios. (3) Patient Documentation \cite{jung2024enhancing}: They can automatically translate patient records, intake forms, discharge summaries, and consent forms, ensuring that medical staff can access and understand crucial information promptly, regardless of the original language

$\bullet$ \textit{Drug Discovery.}
The integration of Large Language Models (LLMs) into drug discovery marks a significant advancement in pharmaceutical research. These AI-driven models, trained on extensive biomedical data, expedite literature review and data mining, uncovering novel drug targets and disease mechanisms. LLMs improve computational drug development by predicting molecular properties such as solubility and toxicity and designing optimized compounds for therapeutic efficacy. They play a crucial role in virtual screening, efficiently identifying promising drug candidates from vast chemical libraries. In addition, LLMs contribute to the design and analysis of clinical trials, optimizing patient recruitment and treatment personalization. As LLMs continue to evolve, they promise to accelerate the discovery of innovative therapies, heralding a new era of precision medicine.

It's crucial to note that while these tools can greatly enhance the efficiency and accuracy of clinicians, they should not replace human expertise and clinical judgment. Med-LLMs are meant to augment rather than replace the role of healthcare providers, ensuring that diagnoses are made with appropriate context and ethical considerations. Direct patient care and final diagnostic decisions should always be under the supervision of qualified medical professionals.

\subsection{Unique Challenges for Med-LLMs} 
The application of Artificial Intelligence (AI) in healthcare encounters a series of unique challenges. These challenges include managing protected health information to ensure data privacy and security, integrating seamlessly with clinical workflows to enhance efficiency, and ensuring the safety and traceability of model deployments to maintain the trust and accountability of patients and medical institutions. In this section, we will delve into the various obstacles encountered by LLMs when applied in the medical field, aiming to provide valuable insights for their further development in the healthcare industry.

$\bullet$ \textit{Protected Health Information.}
In the construction of digital medical infrastructure, implementing stringent security protocols and privacy measures is crucial. Healthcare entities bear significant responsibility to ensure that patients have confidence in the management, transmission, and use of their health data. Furthermore, these entities must adhere to relevant legal and ethical standards to strictly protect the confidentiality of patient information, thereby maintaining the credibility and reliability of medical services~\cite{jawad_security_2024, thirunavukarasu2023large}.

$\bullet$ \textit{Clinical Workflows.} LLMs demonstrate significant potential in areas such as education, laboratory, and pathology analysis~\cite{arvisais-anhalt_establishing_2024}, and facilitating Shared Decision Making (SDM)~\cite{lawson_mclean_large_2024}. However, integrating LLMs into existing hospital workflows presents challenges. This integration process requires meticulous planning to adapt to specific medical environments and to overcome technological, ethical, and operational barriers.
Only through careful planning and effective management of these challenges can LLMs be seamlessly incorporated into hospital workflows, thereby enhancing the efficiency and quality of patient care~\cite{clusmann_future_2023}.

$\bullet$ \textit{Safety and Accountability.} Due to their ``black box'' nature, LLMs present complex challenges in interpretability. Therefore, to ensure the accountability and safety of LLMs, stringent supervision and verification measures must be implemented. These measures are designed to maintain the integrity and credibility of LLM applications, thereby ensuring their effectiveness and reliability in practical use~\cite{thirunavukarasu2023large}.

\section{Trustworthiness and Safety}
In this section, we delve into the challenges pertaining to the trustworthiness and safety of LLMs in medicine and then summarize the countermeasures taken to address these issues. We hope to promote safer and more reliable applications of LLMs in medicine.

$\bullet$ \textit{Fairness.} As a concept stemming from sociology, economics, and law, fairness is described by the Oxford English Dictionary as ``imperfect and just treatment or behavior without favoritism or discrimination"~\cite{li2023survey,guo2023fair}.  In NLP, ensuring fairness requires the elimination of social biases embedded in the model, as these biases can lead to unfair treatment of specific groups~\cite{garrido2021survey} in the process of model coding and task processing.  Therefore, a fair language model must be free from bias, which makes fairness and social bias intertwined in NLP research. Research has revealed biases in healthcare~\cite{nori2023capabilities}, and LLMs are able to capture these biases from training data~\cite{li2023survey, zhang2023chatgpt, chhikara2024few, dwivedi2023breaking} and amplify existing biases~\cite{omiye2023large, bender2021dangers}, producing medically unfair outputs.
Given the various biases (e.g., race, gender, and disability) present in clinical practice and research corpora used for training large language models~\cite{nori2023capabilities, karabacak2023embracing, omiye2024large}, and the potential harms that unfair large language model systems can bring to medicine~\cite{ blease2023chatgpt}, ensuring the fairness of large language models in medical applications is crucial. The medical bias generated by large models can come from skewed data and model limitations.
If biased training data is used, LLMs may retain or even enhance such bias~\cite{mesko2023imperative, thirunavukarasu2023large, omiye2024large}.
At the same time, bias also comes from the limitations of models such as design specifications, structures, and algorithms~\cite{ferrara2023should}. 

At present, there are various technologies for quantifying LLM bias in the medical field. Among them, disparity metrics play an important role in bias detection. Taking BiasMedQA~\cite{schmidgall2024addressing} as an example, it is a benchmark for evaluating cognitive bias in medical tasks, which can effectively quantify the specific levels of cognitive bias in six LLMs (GPT-4, Mixtral-8x70B, GPT-3.5, PaLM-2, Llama 2 70B-chat, and PMC Llama 13B focusing on the medical field).
In addition, counterfactual evaluation can provide insight into the model's understanding of causal relationships by constructing hypothetical scenarios, thereby revealing and quantifying potential biases in the medical field~\cite{chu2023causal,huang2019reducing}. Current strategies to eliminate LLM bias in the medical field are also diversifying.
For open-sourced LLMs, reinforcement learning using feedback from clinicians, etc.~\cite{zack2024assessing} is a promising way to counteract bias.
For non-open source LLMs, it is often necessary to design debiasing programs according to the downstream task training strategy.
The medium-sized LLMs previously developed by GPT-3 use pre-training and fine-tuning paradigms as the main training strategies and such LLMs' social bias can be debiased from a variety of perspectives. For example, the data augmentation technique ~\cite{parray2023chatgpt} is used to optimize the training data during the pre-training phase to effectively reduce bias. LLMs with 100 million parameters are developing rapidly based on demonstrations. Since the representations of most closed-source large-size LLMs are not available, it is more difficult to eliminate bias in the responses of large-scale LLMs, and the commonly used methods include instruction fine-tuning ~\cite{singhal2023large} and prompt engineering ~\cite{schmidgall2024addressing, mathavan2023mitigating} strategies.

$\bullet$ \textit{Accountability}. Lack of accountability in LLMs is recognized as an obstacle hindering its application in the medical field~\cite{solomon2023chatgpt}.
To address this challenge, clinical practice, and research ensure the reliability of LLMs through interpretive methods and human oversight procedures. Interpretive methods increase the transparency of the model and help people understand its decision-making process. \cite{shariatmadari2024harnessing} enhanced the interpretability of LLMs in the biomedical field by integrating knowledge graphs with large language models and visualizing attention probabilities.
In addition to attention visualization, some recent studies~\cite{elsborg2023using} have used Local explanation models to assist LLMs and improve the interpretability of LLMs in the medical field. Human oversight procedures are crucial for ensuring transparency and reliability in LLM operations, encompassing clinical trial protocol development~\cite{wheeler2020clinical} and continuous monitoring of model performance~\cite{mesko2023imperative, karabacak2023embracing}, thereby involving human experts to bridge the accountability gap and improve LLMs' effectiveness in medical scenarios.

$\bullet$ \textit{Privacy.} Data privacy is an important challenge for medical applications. LLMs inadvertently capture sensitive information, leading to privacy leakage during text generation ~\cite{das2024security}. Unintentional data memory, data leakage, and potential leakage of personal information have become the main privacy challenges faced by LLMs~\cite{pan2020privacy}. At present, privacy protection for privacy leakage and passive privacy attacks can be divided into pre-training and fine-tuning protection methods according to the location of privacy protection~\cite{yan2024protecting}. De-identification technology protects privacy by performing thorough data processing of sensitive information during the pre-training phase. DeID-GPTcite~\cite{liu2023deid} uses the powerful Named Entity Recognition (NER) capability of large models to identify sensitive information, to achieve automatic de-identification, showing high accuracy and significant reliability, and is one of the earliest studies using LLM for medical text data processing and de-identification. At the same time, MA Rahman~\cite{rahman2023survey} proposed that federated learning~\cite{zhao2024llm}, differential privacy~\cite{singh2024whispered} can reduce data breaches in the healthcare industry due to LLM API calls during fine-tuning.


$\bullet$ \textit{Robustness}. One of the possible future research directions of Med-LLMs is to explore the construction of effective adversarial test samples in the medical field, including the construction of synthetic anomaly cases ~\cite{yuan2024revisiting} and boundary stress testing ~\cite{wang2024stumbling} to evaluate the robustness of large language models in the medical field. Due to the limited research in this area, especially as Alberts pointed out in the 2023 study ~\cite{alberts2023large}, there are many challenges in constructing medically relevant adversarial test samples, so future research should focus on overcoming these challenges and developing adversarial test methods suitable for the medical field. There have been many studies that use uncertainty quantification methods to improve robustness. For example, Ke Shen et al.~\cite{shen2023formalism} proposed a formalism and method to improve the robustness of large language models by using risk-adjusted confidence scores. Future research can continue to explore how to apply uncertainty quantification techniques to large language models in the medical field to improve the robustness of the model and reduce risks. At the same time, considering the difficulties of large language models in processing out-of-distribution inputs, future research can also focus on how to use techniques similar to LLM-TTA ~\cite{o2024improving} to improve the robustness of the model to unknown or rare cases in the medical field. By effectively using the augmentation data generated by the model as a means of augmentation at test time, we can reduce the dependence of the model on expensive augmentation while maintaining the performance of the model, which will have important implications for advancing the application of large language models in the medical field.

\section{Future Directions}
\label{future}
Based on the comprehensive survey of medical large language models presented in this paper, several promising future directions can be identified to further advance the field. These directions encompass algorithmic advancements, industrial transformations, and policy developments.

One key area for future research is the exploration of novel algorithmic approaches to enhance the capabilities of Med-LLMs. The integration of multimodal learning holds significant potential, where LLMs can be trained to process and understand various types of medical data, such as imaging, sensor readings, and genetic information. This holistic approach to data integration can lead to more comprehensive and accurate medical insights, enabling LLMs to provide a more complete picture of a patient's health status. Furthermore, the emerging field of robot-assisted learning presents exciting opportunities for Med-LLMs. By leveraging embodied cognition and human-AI collaboration, LLMs can be integrated with medical robotics systems to perform complex surgical procedures and assist in patient care. This synergistic combination of natural language processing and physical embodiment can revolutionize the way medical procedures are performed and improve patient outcomes.

In addition to algorithmic advancements, the future of Med-LLMs is expected to drive significant transformations in the healthcare industry. One major area of impact is preventative and precision medicine. By harnessing the predictive power of LLMs, early disease detection and personalized care plans can be developed for individual patients. LLMs can analyze vast amounts of patient data, including electronic health records, genetic information, and lifestyle factors, to identify risk factors and predict potential health issues. This proactive approach to healthcare has the potential to improve patient outcomes, reduce healthcare costs, and promote a shift towards preventative care. Moreover, the integration of LLMs into clinical workflows can streamline various processes, such as medical documentation, diagnostics, and decision support. LLMs can assist healthcare professionals by automatically generating clinical notes, providing evidence-based recommendations, and issuing alerts for potential adverse events. This can greatly reduce the administrative burden on healthcare providers and allow them to focus on delivering high-quality patient care. Another promising application of Med-LLMs is in drug discovery. By analyzing vast amounts of biomedical literature and molecular data, LLMs can identify promising compounds and optimize molecular designs, accelerating the development of novel therapies. LLMs can uncover hidden patterns and relationships in drug-target interactions, predict potential side effects, and suggest novel drug combinations. This can significantly reduce the time and cost associated with drug discovery and development, leading to faster translation of research into clinical applications.

To fully realize the potential of Med-LLMs, supportive policy frameworks need to be developed. Regulatory frameworks for AI in healthcare are crucial to ensure the safety, efficacy, and ethical deployment of these technologies. Clear guidelines and standards for the development, validation, and monitoring of Med-LLMs should be established to build trust among healthcare providers and patients. These frameworks should address issues such as data privacy, algorithmic bias, and transparency in decision-making. Additionally, reimbursement models and incentives should be designed to encourage the adoption and responsible use of LLMs in clinical practice. Policymakers should work closely with healthcare stakeholders to create an enabling environment that fosters innovation while prioritizing patient safety and privacy.

Therefore, the future of Med-LLMs holds immense promise for revolutionizing healthcare delivery and improving patient outcomes. By advancing algorithmic capabilities, driving industrial transformations, and developing supportive policy frameworks, the medical community can harness the full potential of LLMs. However, it is crucial to approach this future with a balanced perspective, acknowledging the current limitations and challenges that need to be addressed. Open problems, such as robustness, interpretability, and data constraints, require further research and development. Adoption challenges, including trust, usability, and integration into existing workflows, must be carefully navigated to ensure the successful deployment of LLMs in clinical practice. As the field continues to evolve, close collaboration between researchers, healthcare professionals, policymakers, and industry stakeholders will be essential to shape a future where Med-LLMs can truly transform healthcare for the benefit of patients worldwide.

\section{Conclusions}
In conclusion, this comprehensive survey has provided a detailed overview of the rapidly evolving field of medical large language models, exploring their fundamental architectures, progressive enhancements, and trustworthy considerations. The paper has traced the development history of LLMs, discussed the transition from general-purpose to domain-specific models, and highlighted their significant potential in revolutionizing various aspects of healthcare. It has examined key algorithmic advancements, such as clinical reasoning, knowledge graph integration, and retrieval-augmented generation, which enhance LLMs' performance and usability in handling complex medical queries and generating accurate responses. The survey has also explored the diverse applications of LLMs, ranging from clinical documentation and diagnosis assistance to patient communication and medical education, demonstrating their immense potential in streamlining healthcare processes and improving patient outcomes. In addition, the paper has emphasized the critical importance of trustworthy and safe deployment of Med-LLMs, discussing challenges associated with fairness, accountability, privacy, and robustness, and highlighting the need for rigorous evaluation, ethical considerations, and regulatory frameworks. Looking toward the future, the survey has identified promising research directions, including algorithmic advancements, industrial transformations, and policy developments to support the responsible growth of Med-LLMs.

This comprehensive survey has provided a valuable resource for researchers, practitioners, and stakeholders interested in the field of Med-LLMs. It has highlighted the immense potential of these models in revolutionizing healthcare delivery while also emphasizing the need for responsible development and deployment. As the field continues to evolve, close collaboration between researchers, healthcare professionals, policymakers, and industry stakeholders will be essential to harness the full potential of Med-LLMs and shape a future where they can truly transform healthcare for the benefit of patients worldwide.

\bibliographystyle{IEEEtran}
\bibliography{IEEEtran}


\vfill

\end{document}